\documentclass[journal]{IEEEtran}
\usepackage{pgfplots}
\usepackage{multicol}
\usepackage[colorlinks=true]{hyperref}

\usepackage{bm}
\usepackage{url}
\usepackage{graphicx} 
\usepackage{mathptmx} 
\usepackage[mathscr]{euscript}  
\usepackage{amsmath} 
\usepackage{amssymb}  
\usepackage{flushend}
\usepackage{siunitx}
\usepackage{xcolor}
\usepackage{adjustbox}
\usepackage{array}
\usepackage{multirow}
\usepackage{booktabs}

\usepackage[inline]{enumitem}
\usepackage{microtype}
\usepackage{amsmath}
\usepackage{threeparttable}
\usepackage{caption}
\usepackage[resetlabels]{multibib}

\newcommand{\irow}[1]{
  \begin{bmatrix}#1\end{bmatrix}%
}

\usepgfplotslibrary{groupplots}
\pgfplotsset{compat=1.13}

\newcommand{\secref}[1]{Sec.~\ref{#1}}
\renewcommand{\eqref}[1]{Eqn.~(\ref{#1})}
\newcommand{\figref}[1]{Fig.~\ref{#1}}
\newcommand{\tabref}[1]{Tab.~\ref{#1}}

\newcolumntype{P}[1]{>{\centering\arraybackslash}p{#1}}
\pdfoutput=1

\begin{document}

\title{Dynamic Object Removal and Spatio-Temporal RGB-D Inpainting via\\Geometry-Aware Adversarial Learning}
\author{Borna Be\v{s}i\'{c},
        and~Abhinav Valada
\thanks{Department of Computer Science, University of Freiburg, Germany}}

%

\markboth{\copyright~2022 IEEE}
\IEEEaftertitletext{\vspace{-1\baselineskip}}


\maketitle

\begin{abstract}
Dynamic objects have a significant impact on the robot's perception of the environment which degrades the performance of essential tasks such as localization and mapping. In this work, we address this problem by synthesizing plausible color, texture and geometry in regions occluded by dynamic objects. We propose the novel geometry-aware DynaFill architecture that follows a coarse-to-fine topology and incorporates our gated reccurrent feedback mechanism to adaptively fuse information from previous timesteps. We optimize our architecture using adversarial training to synthesize fine realistic textures which enables it to hallucinate color and depth structure in occluded regions online in a spatially and temporally coherent manner, without relying on future frame information. Casting our inpainting problem as an image-to-image translation task, our model also corrects regions correlated with the presence of dynamic objects in the scene, such as shadows or reflections. We introduce a large-scale hyperrealistic dataset with \mbox{RGB-D} images, semantic segmentation labels, camera poses as well as groundtruth \mbox{RGB-D} information of occluded regions. Extensive quantitative and qualitative evaluations show that our approach achieves state-of-the-art performance, even in challenging weather conditions. Furthermore, we present results for retrieval-based visual localization with the synthesized images that demonstrate the utility of our approach.
\end{abstract}


%
\IEEEpeerreviewmaketitle

\section{Introduction}

\IEEEPARstart{N}{avigation} in urban environments pose a significant challenge for autonomous robots due to the sheer number of dynamic objects (\textit{e.g.} pedestrians, vehicles, cyclists)
A dynamic object, in this context is any movable object that is currently moving through a scene or has an ability to do so. Such objects continually occlude the scene which hinders essential tasks such localization, mapping, and reasoning. Several solutions have been proposed to tackle this problem from filtering out regions that contain dynamic objects~\cite{sun2017improving, boniardi2019robot} to assuming a static scene and classifying dynamic object regions as outliers~\cite{klein2007parallel, valada2018incorporating}. More recently, learning-based methods~\cite{bescos2019empty, yu2018free, pathakCVPR16context} have shown promising results by inpainting dynamic object regions in images, with the background structure behind them. These methods first detect regions containing dynamic objects at the pixel-level using semantic segmentation~\cite{bescos2019empty} or motion segmentation~\cite{Vertens17Icra}, followed by synthesizing the background in those regions using an encoder-decoder architecture. Moreover, there are numerous other applications to this inpainting task such as photo editing, video restoration, augmented reality and diminished reality, which makes it a widely studied and fundamental task.

Classical computer vision methods typically fall short in producing visually appealing results as they often include over-smoothed content with lack of textures, content that does not match the semantic context and geometry of an occluded area. Whereas, learning-based methods leverage experience and learn semantic priors from large amounts of examples which yields spatially consistent results. However, these methods still fail at recovering the geometry and yields severe temporal artifacts when applied to a sequence of images. Video inpainting methods aim to address the latter by introducing an additional constraint of temporal coherency. These methods often leverage optical flow taking advantage of the fact that inpainting information comes not only from the neighborhood of the target region in the current frame but also propagates from the context defined by both past and future frames. However, they make several assumptions that are often violated in real-world online applications such as robotics, for example, restrictions on the motion of the camera, notion of visibility of the occluded region in past and future frames, illumination changes, perspective changes of dynamic objects, among others. Moreover, as both image and video inpainting methods only aim to complete missing regions, they often leave behind artifacts induced by dynamic objects, such as shadows or reflections, which degrades the performance of certain tasks or yield unappealing results.\looseness=-1

\begin{figure}
\footnotesize
\centering
\includegraphics[width=\linewidth]{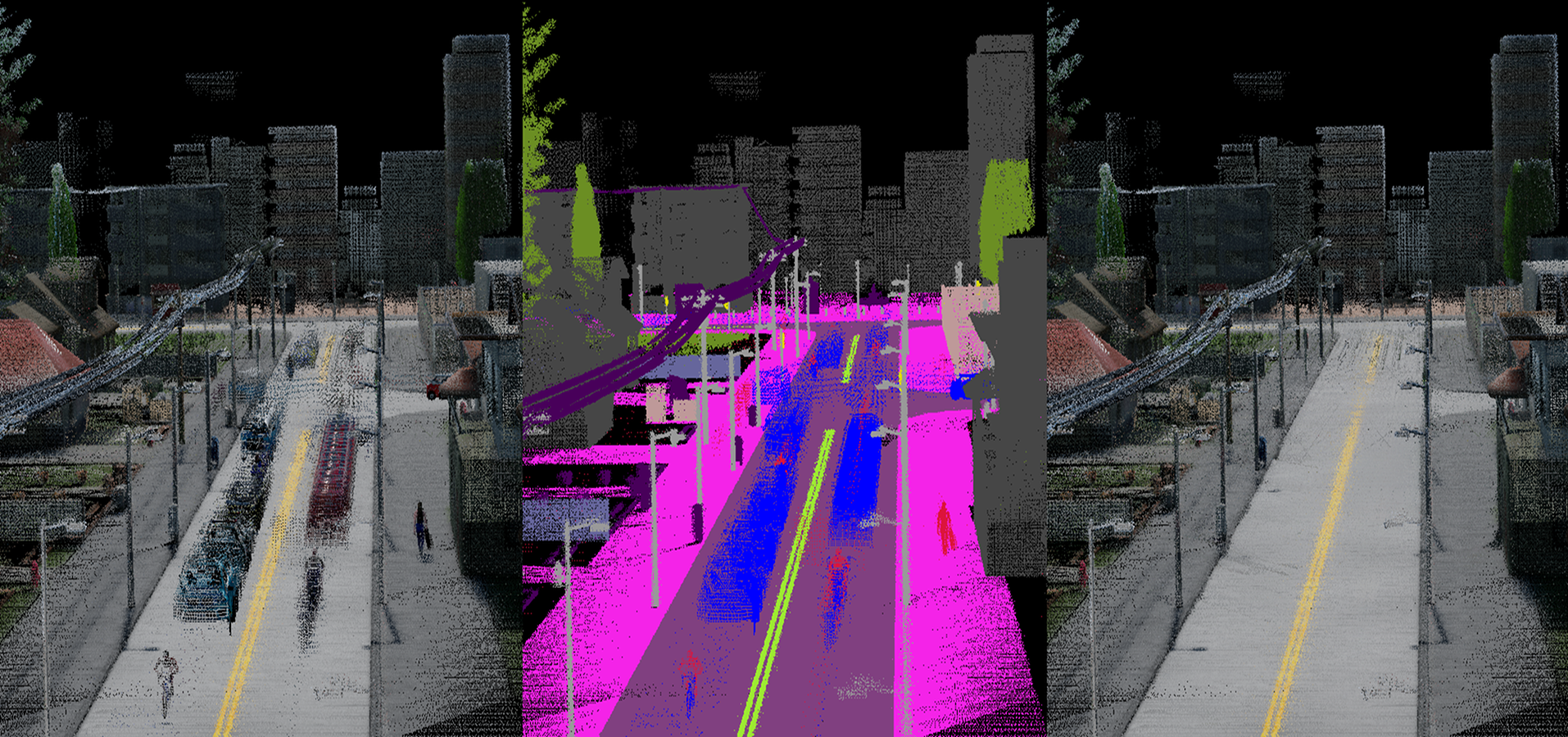}
\caption{Point cloud visualization showing the dynamic object removal and inpainting using our proposed DynaFill model. The point cloud built from: input RGB-D frames (left), semantic segmentation output with dynamic object masks (center), and the inpainted RGB-D output (right).}
\label{fig:coverfig}
\end{figure}

In this paper, we propose the novel geometry-aware DynaFill architecture that synthesizes parts of a scene occluded by dynamic objects with plausible color, texture and geometry (\figref{fig:coverfig}) from a stream of RGB-D images, while providing effective solutions to the aforementioned problems. In fact, our method is agnostic towards a method of detecting or segmenting objects, and therefore requires only a mask that indicates the pixels to be inpainted. Our inpainting architecture follows a coarse-to-fine topology and consists of three sub-networks: coarse inpainting, refinement image-to-image translation, and depth completion streams. We employ a semantic segmentation stream to identify dynamic object regions as a mask extraction front-end for our inpainting system. First we inpaint the regions on a coarse scale using the coarse inpainting subnetwork. Subsequently, the refinement image-to-image translation stream trained in an adversarial manner, takes the coarsely inpainted image as input and adds spatially consistent fine details while removing any artifacts caused by dynamic objects such as shadows or reflections. The depth completion stream then regresses depth values in occluded regions, conditioned on the inpainted RGB image. We propose a recurrent gated feedback mechanism that adaptively selects relevant information from the previously inpainted image and fuses them into the refinement image-to-image translation network to enforce temporal consistency. By training our entire inpainting network in an end-to-end manner and by conditioning the depth completion using the inpainted image as well as utilizing the previously inpainted depth map in the recurrent feedback, we allow the image and depth sub-networks to supervise each other. The inpainted depth information, therefore, allows us to reason about the geometry of the scene. By employing the inpainted depth in the process of forward warping of previous inpainted RGB frames, we achieve a joint level of supervision which makes our approach geometry-aware.

To the best of our knowledge, the DynaFill model is the first spatially and temporally consistent RGB-D inpainting approach that does not rely on future frame information.\looseness=-1

To facilitate this work, we introduce a first-of-a-kind large-scale hyperrealistic dataset of urban driving scenes that contains paired RGB-D images with groundtruth information of occluded regions, semantic segmentation labels, and camera pose information. Our dataset consists of a large number of dynamic objects and weather conditions that make spatio-temporal inpainting extremely challenging. We perform extensive quantitative and qualitative comparisons with both image inpainting as well as video inpainting methods that demonstrate that DynaFill achieves state-of-the-art performance while being faster than other video inpainting methods. Additionally, we present retrieval-based visual localization experiments using the synthesized images that show a substantial improvement in localization performance.

In summary, the following are the main contributions:
\begin{enumerate}[noitemsep]
    \item The novel DynaFill architecture that consists of a semantic segmentation sub-network, a coarse inpainting sub-network, a refinement image-to-image translation sub-network, and a depth inpainting sub-network.
    We employ a gated recurrent feedback mechanism that enforces temporal consistency by adaptively selecting and fusing relevant information from previously inpainted images.
    \item A mechanism that enables the image and depth inpainting sub-networks to supervise each other. The joint training of the video inpainting network that uses the output of the depth completion network improves the performance of both RGB inpainting and depth inpainting. It makes our approach the first spatially and temporally consistent RGB-D video inpainting method that does not require future frame information.  
    \item We introduced a large-scale driving dataset for learning to inpaint dynamic object regions in RGB-D videos. Our dataset is the first to contain groundtruth information of occluded regions in RGB-D driving videos.
    \item Extensive quantitative and qualitative evaluations of our proposed model, complemented with detailed ablation studies and explainable visualizations, and experiments that demonstrate the utility of our approach as an out-of-the-box front end for location and mapping systems.
    \item We make the code and trained models publicly available at \url{http://rl.uni-freiburg.de/research/rgbd-inpainting}.
\end{enumerate}

\section{Related Work}


{\parskip=5pt
\noindent\textbf{Exemplar-based inpainting}: These methods fill target holes using texture statistics from adjacent known regions. Thy typically use various iterative diffusion-based~\cite{bertalmio2000image, ballester2001filling} or patch-based techniques~\cite{huang2014image, barnes2009patchmatch, Darabi12:ImageMelding12}. Although these methods yield visually appealing results, they are not suitable for filling large holes due to their inability to preserve structure and their large runtimes make them unsuitable for real-time applications. On the other hand, fast inpainting methods~\cite{journals/jgtools/Telea04} that trade off quality for speed produce blurry content with lack of texture and is geometrically inconsistent.}

{\parskip=5pt
\noindent\textbf{Learning-based Image Inpainting}: In recent years, CNN-based methods have significantly outperformed earlier works, both in visual quality and runtime. The introduction of Generative Adversarial Networks (GANs)~\cite{NIPS2014_5423} has transformed learning-based image inpainting by casting it into a conditional image generation task. Pathak~\textit{et~al.} propose Context Encoders~\cite{pathakCVPR16context} that employ GAN loss along with pixel-wise reconstruction loss to generate contents of an arbitrary image region conditioned on its surroundings. Most of the initial learning-based methods were limited to inpainting a single target region of rectangular shape~\cite{pathakCVPR16context, Yang_2017_CVPR, yu2018generative}. Iizuka~\textit{et~al.}~\cite{IizukaSIGGRAPH2017} and Yu~\textit{et~al.}~\cite{yu2018free} were the first to tackle the challenge of free-form image inpainting of arbitrary number of regions. The CM model~\cite{IizukaSIGGRAPH2017} builds upon~\cite{pathakCVPR16context} by incorporating a global discriminator that considers the entire image to assess if the inpainting is coherent as a whole and a local discriminator that only considers a small area centered at the completed region to ensure local consistency. DeepFill~v2~\cite{yu2018free} builds upon partial convolutions~\cite{Liu_2018_ECCV} and contextual attention~\cite{yu2018generative} by introducing learnable gated convolutions together with SN-PatchGAN, which alleviates the need for two different discriminators and substantially stabilizes the training.}

More recently, several methods~\cite{nazeri2019edgeconnect, UlyanovVL17} have been introduced to incorporate more prior knowledge into the inpainting task. Nazeri~\textit{et~al.} propose EdgeConnect~\cite{nazeri2019edgeconnect}, a two-stage adversarial model in which the first network hallucinates missing edges in target regions, while the second network performs inpainting conditioned on the synthesized edges. Deep Image Prior~\cite{UlyanovVL17} demonstrates that the convolutional structure of a network is sufficient to capture a significant amount of low-level image statistics for inpainting. In contrast, Bescos~\textit{et~al.}~\cite{bescos2019empty} formulate the problem as an image-to-image translation task which enables them to fill holes coarsely and correct regions in the scene that are correlated with the presence of dynamic objects.\looseness=-1

\begin{figure*}[h]
\centering
\includegraphics[width=\textwidth]{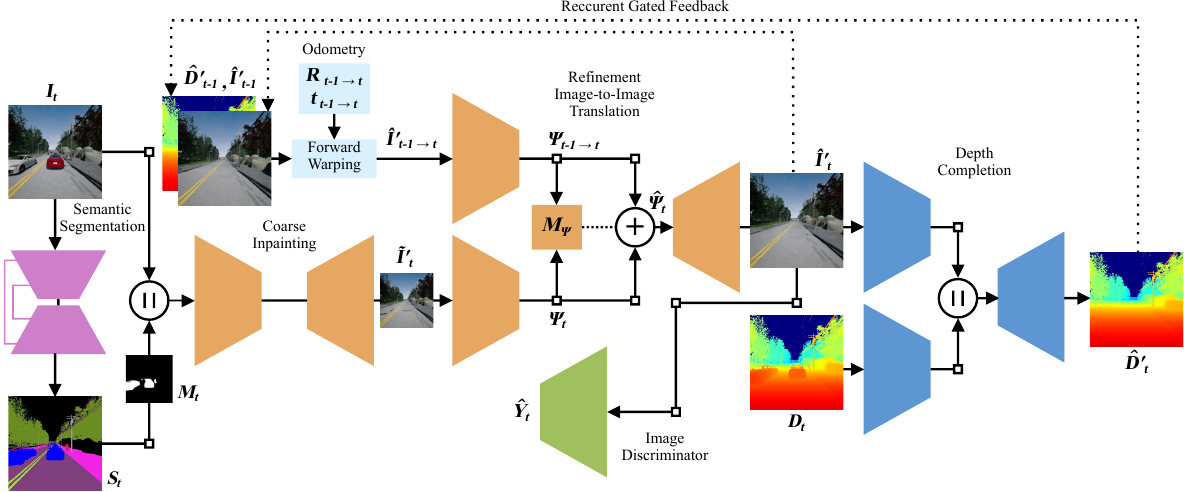}
\caption{Schematic representation of our DynaFill architecture. The image $I_t$ is first coarsely inpainted based on the spatial context in regions occluded by dynamic objects $M_t$, which is obtained from the semantic segmentation mask $S_t$. Subsequently, the inpainted image from the previous timestep $\hat{I'}_{t-1}$ is warped into the current timestep using odometry and the inpainted depth map $\hat{D'}_{t-1}$ in our recurrent gated feedback mechanism. The coarsely inpainted image $\widetilde{I'}_{t}$ and the warped image $\hat{I'}_{t-1\,\rightarrow\,t}$ are then input to the refinement stream that fuses feature maps through a gating network using the learned mask $\bm{M_{\Psi}}$. An image discriminator is employed to train the network in an adversarial manner to yield the final inpainted image $\hat{I'}_{t}$. Simultaneously, the depth completion network fills the regions containing dynamic objects in the depth map $D_t$, conditioned on the inpainted image $\hat{I'}_{t}$.}
\label{fig:pipeline}
\end{figure*}

{\parskip=5pt
\noindent\textbf{Learning-based Video Inpainting}: A complementary class of methods address the challenge of inpainting temporal image sequences, typically by formulating it as a learning-based video inpainting task. Kim~\textit{et~al.}~\cite{kim2019deep} model video inpainting as a sequential multi-to-single frame inpainting problem to gather features from neighbor frames and synthesize missing content based on them. Woo~\textit{et~al.}~\cite{woo2019alignandattend} tackle the limitation of small temporal window sizes in existing approaches and propose the align-and-attend network to alleviate this problem. Copy-and-Paste Networks~\cite{lee2019cpnet} adopts a similar approach in which a context matching module is used as an attention mechanism to combine the target frame with past and future reference frames aligned through a learned affine transformation. As opposed to directly inpainting temporal frames, Xu~\textit{et~al.}~\cite{Xu_2019_CVPR} first synthesize missing optical flow which is then used to propagate neighboring pixels to fill missing regions. Chang~\textit{et~al.}~\cite{chang2019learnable} present the learnable gated temporal shift module which is incorporated in both the generator and discriminator networks to automatically learn to shift frames temporally to its neighbors. Building upon~\cite{UlyanovVL17}, Zhang~\textit{et~al.}~\cite{zhang2019internal} propose an approach to predict both image frames and optical flow maps for video inpainting by optimizing the network directly on the input video.}\looseness=-1

While the aforementioned prior work have made significant contributions, they still do not address the problem of spatially and temporally coherent RGB-D inpainting when future frames are not available. Since this is a critical requirement for real-world online applications, we propose the novel geometry-aware DynaFill learning framework for removing dynamic objects and inpainting both color as well as depth structure. Our proposed coarse-to-fine network adaptively exploits information from the current and previously inpainted regions using our gated recurrent feedback mechanism to achieve temporal coherence. As opposed to existing work, incorporating the inpainted image for the depth completion and utilizing the previously filled depth map for image inpainting, enables our method to achieve geometrically consistent results from end-to-end optimization of our inpainting architecture. Moreover, our model yields spatio-temporally coherent and visually congruous results by performing both temporal inpainting and image-to-image translation.


Concurrently to this work, a few other approaches have been proposed. Zhang~\textit{et~al.}~\cite{nonlocal} introduces a novel two-step algorithm for image completion. After triangulation-based linear interpolation of available pixels, similar non-local patches grouped as a tensors are completed using a solver under the alternating direction method of multiplier (ADMM) framework. Another interesting recent approach is DSNet~\cite{dynamic-wang} that distinguishes the corrupted region from the valid ones throughout the entire network architecture, and generates plausible content by using two novel deep learning modules, namely, the validness migratable convolution (VMC) and regional composite normalization (RCN) modules.

\section{Technical approach}
\label{sec:technical}

The goal of our approach is to remove dynamic objects from an online stream of RGB-D images, while synthesizing plausible color, texture and geometry in the occluded regions. We aim to inpaint regions of any object that is moving or has a potential to significantly affect the perception of the scene by background (\textit{e.g.} parked cars). For brevity, we call these objects as \textit{dynamic} in the rest of this paper.
Let $z_t = \left(\bm{I}_{t},\ \bm{D}_{t},\ \bm{M}_{t}\right)$ be the current observed frame at the timestep $t$ that potentially contains dynamic objects, where $\bm{I}_{t} \in \mathbb{R}^{3 \times H \times W}$ and $\bm{D}_{t} \in \mathbb{R}^{H \times W}$ denote an image and the corresponding depth map respectively, both of height $H$ pixels and width $W$ pixels. A binary mask $\bm{M}_{t} \in \{0, 1\}^{H \times W}$ indicates which pixels belong to dynamic objects such as cars, trucks, pedestrians, and cyclists. Let $x_t = (\bm{I}'_t,\ \bm{D}'_t)$ be its corresponding frame that does not contain any dynamic objects. We define our task as observing an input RGB-D stream of a dynamic environment and transforming it into the equivalent RGB-D stream of a static environment. Assuming such a transformation function $f$ exists, the conditional probability distribution for a single time step can be written as
\begin{equation*}
    p\left[x_t\ \middle|\ z_{1:t},\ x_{1:t-1}\right] = p\left[f(z_{1:t},\ x_{1:t-1})\ \middle|\ z_{1:t},\ x_{1:t-1}\right].
\end{equation*}

To reduce the complexity of modelling $f$ and to make the computation feasible, we follow the approach of Wang~\textit{et~al.}~\cite{wang2018video} and assume that the Markov property holds. By making the Markov assumption of $L$-th order, where the current output depends only on the last $L$ outputs and the current observation, we obtain
\begin{equation*}
    p\left[x_t\ \middle|\ z_t,\ x_{t-L:t-1}\right] = p\left[f(z_t,\ x_{t-L:t-1})\ \middle|\ z_t,\ x_{t-L:t-1}\right].
\end{equation*}

This allows us to factorize the conditional probability distribution for the whole stream as
\begin{equation*}
    p\left[x_{1:t}\ \middle|\ z_{1:t},\ x_{1:t-1}\right] = \prod_{i=1}^{t} p\left[f(z_i,\ x_{i-L:i-1})\ \middle|\ z_i,\ x_{i-L:i-1}\right].
\end{equation*}

\begin{figure*}
    \centering
    \footnotesize
    \setlength{\tabcolsep}{0.5cm}
    {\renewcommand{\arraystretch}{0.5}
    \begin{tabular}{P{8cm}P{8cm}}
    \includegraphics[width=\linewidth]{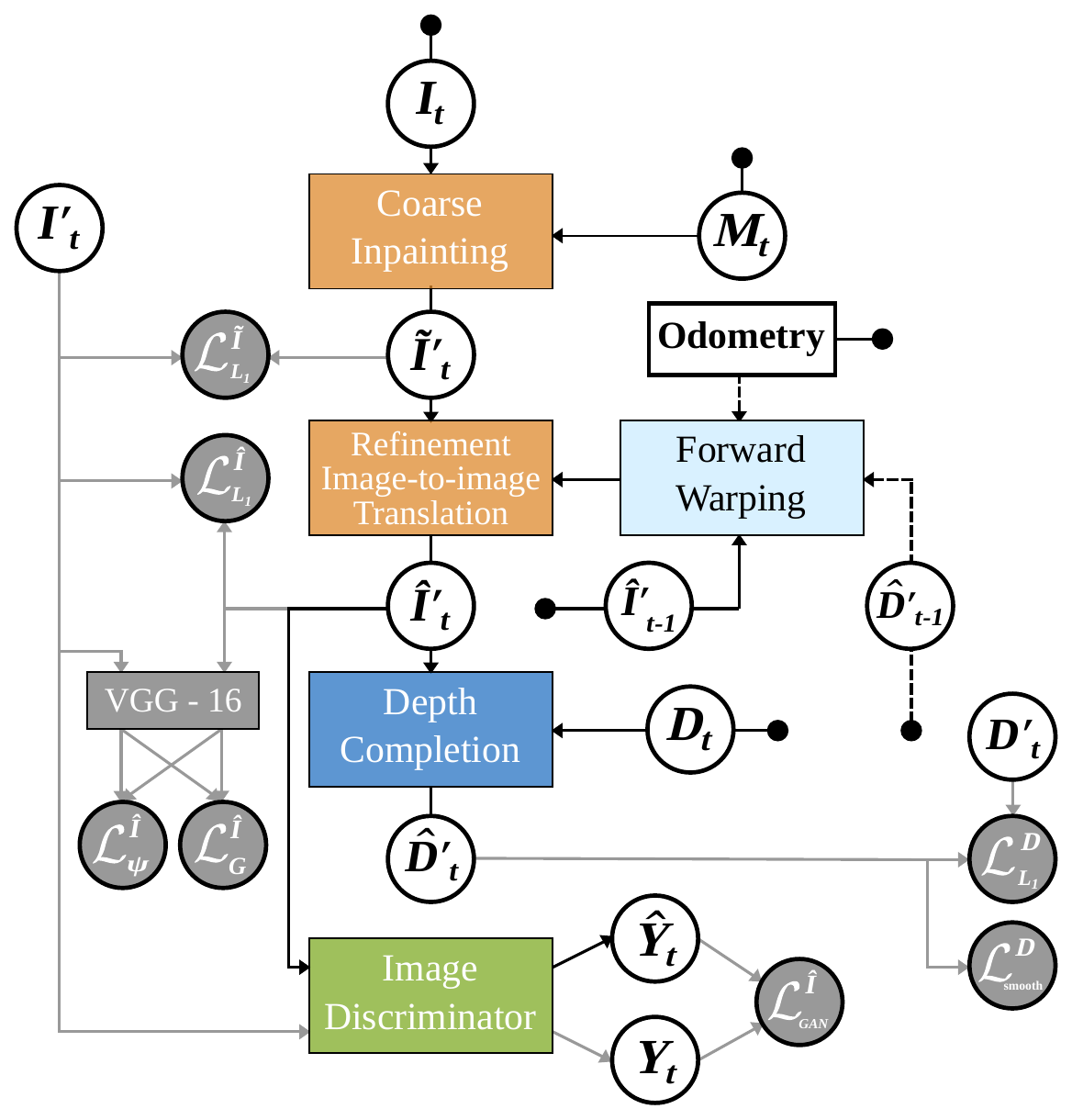} &         \includegraphics[width=0.95\linewidth]{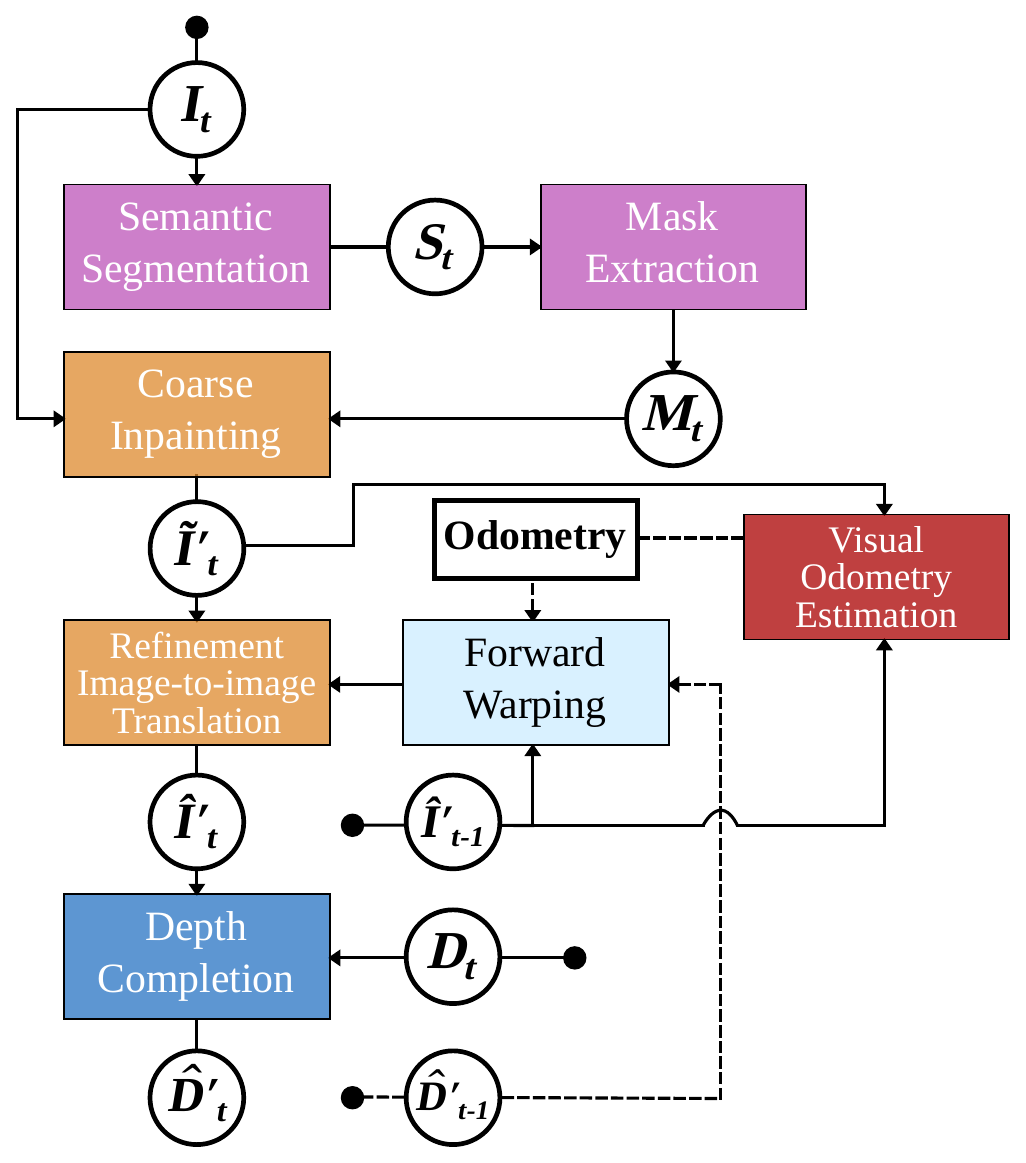} \\
    \\
    (a) Training Phase & (b) Testing Phase \\
    \end{tabular}}
    \caption{Illustration of the computational graphs of our DynaFill model during the (a) training phase together with loss functions, and the (b) testing phase.}
    \label{fig:train_inference}
\end{figure*}

Our approach is to model the underlying function $f$ by learning a function $\hat{f}(z_t,\ \hat{x}_{t-L:t-1}) = \hat{x}_t$ such that the learned conditional probability distribution matches the original conditional probability distribution. We represent $\hat{f}$ with a feed-forward deep neural network that operates in a recurrent manner. More specifically, our DynaFill architecture consists of four sub-networks: semantic segmentation, coarse inpainting, refinement image-to-image translation, and depth completion. An overview of our framework is shown in \figref{fig:pipeline} and block diagrams in the form of computational graph nodes during the training and testing phase are shown in \figref{fig:train_inference}.

\subsection{Semantic Segmentation and Coarse Inpainting}
\label{sec:coarse_inpainting}

The first two sub-networks in our architecture identify the pixels that belong to dynamic objects and coarsely inpaint the occluded regions.
For obtaining a binary mask that indicates regions of objects to inpaint, we employ our AdapNet++~\cite{valada19ijcv} semantic segmentation network that we separately pre-train on our hyperrealistic dataset with the same training protocol as in the original work. Note that any other method that provides a binary mask of object regions can be directly used as a replacement. Removing objects based on their semantic class is useful in a mapping scenario where we want to obtain a clean static map of the environment, or in cases of photo editing where a coherent synthesis of the background is desired. Furthermore, motion segmentation approaches, such as our previously proposed SMSnet~\cite{Vertens17Icra}, can be employed for localization or SLAM where only currently moving objects should be considered. In this work, we aim to remove all the occurrences of certain semantic object classes that could be moving. Therefore, the prediction of the semantic segmentation network is used as an approximation for identifying dynamic objects, which can produce false negatives for classes on which the network was not trained on.

We first pass the image from the current timestep $\bm{I}_{t}$ through the semantic segmentation network to obtain a semantic mask $\bm{S}_{t} \in \{0, 1, ..., n_c\}^{H \times W}$, where $n_c$ is the number of semantic classes. We then extract a binary target region mask $\bm{M}_{t}$ from the semantic mask. Subsequently, we concatenate $\bm{I}_{t} \odot \left(\bm{1}_{\scriptscriptstyle H,W} - \bm{M}_{t}\right)$ with $\bm{M}_{t}$ along the channel dimension and feed it into the coarse inpainting network which has a fully convolutional encoder-decoder topology. Here, $\bm{I}_{t} \odot \left(\bm{1}_{\scriptscriptstyle H,W} - \bm{M}_{t}\right)$ ignores regions that contain dynamic objects by setting the values of those pixels to zero, $\odot$ denotes the Hadamard product and $\bm{1}_{\scriptscriptstyle H,W} \in \{1\}^{H \times W}$ denotes a matrix of ones. The encoder of the coarse inpainting network is built upon the ResNet-50~\cite{DBLP:conf/eccv/HeZRS16} architecture with pre-activation residual units and the decoder consists of three upsampling stages where in each stage we perform bilinear upsampling by a factor of two, followed by a $3 \times 3$ convolution with stride 1 and pad 1. We progressively halve the number of channels in each of the convolutions in the decoder and employ a $1 \times1$ convolution with stride 1 in the end to reduce the number of channels to 3. The final coarse inpainting result $ \bm{\widetilde{I}'}_{t} $ is obtained by combining the masked part of the output of the coarse inpainting network $ \widetilde{f} $, and the unmasked part of its input $ \bm{I}_{t} $, as
\begin{equation}
    \small
    \bm{\widetilde{I}'}_{t} = \bm{I}_{t} \odot \left(\bm{1}_{\scriptscriptstyle H,W} - \bm{M}_{t}\right) + \widetilde{f}\left[\bm{I}_{t} \odot \left(\bm{1}_{\scriptscriptstyle H,W} - \bm{M}_{t}\right), \bm{M}_{t}\right] \odot \bm{M}_{t},
\end{equation}
where the values in the input and output image are in the range of $[-1, 1]$. We optimize the network by minimizing the $L_1$ distance between the predicted pixels in the target region $ \bm{\widetilde{I}'}_{t} $ and the corresponding groundtruth pixels $ \bm{I'}_{t} $ through loss function $ \mathcal{L}^{\bm{\widetilde{I}}}$ given by
\begin{equation} \label{eq:l1_coarse}
\mathcal{L}^{\bm{\widetilde{I}}} = \mathcal{L}^{\bm{\widetilde{I}}}_{L_1} = \left\lVert \left(\bm{\widetilde{I}'}_{t} - \bm{I'}_{t}\right) \odot \bm{M}_t \right\rVert_1.
\end{equation}
where $\bm{\widetilde{I}'}_{\,t}$ is a coarsely inpainted image, $\bm{I'}_{\,t}$ is the corresponding groundtruth image without dynamic objects and $\bm{M}_t$ is a mask indicating target regions, i.e. pixels belonging to dynamic objects that need to be inpainted.


\subsection{Refinement Image-to-image Translation}
\label{sec:refinement}

We employ the refinement image-to-image translation sub-network to add fine details to the target region as well as to correct surrounding areas that may contain shadows or reflections. Since we remove all dynamic objects from the image, any present optical flow can only be induced by the ego-motion of the camera. Therefore, we propose a recurrent gated feedback mechanism in which we use odometry data coupled with the inpainted depth map to warp the inpainted image from the previous timestep into the current frame to induce temporal context. We define the relative transformation between the poses of two frames at timesteps $t_1$ and $t_2$ in the form of a rotation matrix $\bm{R}_{t_1\rightarrow t_2} \in \mathbb{R}^{3 \times 3}$ and a translation vector $\bm{t}_{t_1\rightarrow t_2} \in \mathbb{R}^{3}$. Using the depth map $\bm{D}_{t_1}$, a pixel $(u, v)$ of the inpainted image at timestep $t_1$ can be transformed to the new homogeneous coordinates $(u', v', w')$ at timestep $t_2$ as
\begin{equation}
    \begin{bmatrix}
    u' \\
    v' \\
    w'
    \end{bmatrix}
    = \bm{K} \bm{R}_{t_1\rightarrow t_2} \bm{K}^{-1}
    \begin{bmatrix}
    u \\
    v \\
    1
    \end{bmatrix}
    + \bm{K} \frac{\bm{t}_{t_1\rightarrow t_2}}{\bm{D}_{t_1}\left(u,\ v\right)},
\end{equation}
where $\bm{K} \in \mathbb{R}^{3 \times 3}$ is the intrinsic camera matrix and $\bm{D}_{t_1}\left(u,\ v\right)$ is the depth. Note that the valid new image coordinates in the two dimensional Euclidean space is obtained by dividing the homogeneous coordinates by $w'$.
The result of the warping operation is an image $\bm{I}_{t_1\rightarrow t_2}$ (approximation of $\bm{I}_{t_2}$) and a mask $\bm{M}_{t_1\rightarrow t_2}$ indicating the pixels at timestep $t_2$ that have been warped from timestep $t_1$ and are not occluded.

In order to refine the coarsely inpainted image, we employ the same building blocks described in \secref{sec:coarse_inpainting}. The sub-network consists of two encoders that take the coarsely inpainted image $\bm{\widetilde{I}'}_{t}$ and the inpainted image from the previous timestep that has been warped into the current timestep $\bm{\hat{I'}}_{t-1\rightarrow t}$ as input (shown in \figref{fig:pipeline}). That is, during the training the previously inpainted RGB image $\bm{\hat{I'}}_{t-1}$ is warped forward using the ground truth odometry and the previously inpainted depth $\bm{\hat{D'}}_{t-1}$. At inference time, we use SVO~2.0~\cite{7782863} for estimating the odometry. Since our architecture predicts an inpainted RGB frame at each time step, we can use this information to reduce the error induced by dynamic objects in the odometry estimation. For each time step, we use best estimates of two consecutive inpainted RGB frames. That is, we estimate the odometry between steps $t - 1$ and $t$ by using previous refined static RGB frame $\bm{\hat{I}'}_{t-1}$ and the current coarsely inpainted RGB frame $\bm{\widetilde{I}'}_{t}$. This is feasible since the odometry is only required in the refinement sub-network.

We then employ a gating module that takes the output feature maps of the two encoders $\bm{\Psi_{t}}$ and $\bm{\Psi_{t-1\rightarrow t}}$ concatenated along the channels as input and learns a mask $\bm{M_{\Psi}}$. The gating module consists of five $3\times 3$ convolutions with stride 1 and padding 1, each halving the number of channels. Subsequently, we add a $1\times 1$ convolution that yields a single-channel $\bm{M_{\Psi}}$ mask. The output of the gating module is then used compute the fused feature maps as
\begin{equation}
    \bm{\hat{\Psi}_{t}} = \bm{M_{\Psi}} \odot \bm{\Psi_{t}} + \left(1 - \bm{M_{\Psi}}\right) \odot \bm{\Psi_{t-1\rightarrow t}}.
\end{equation}

Finally the resulting fused feature map is fed into the decoder which yields the refined inpainted image $\bm{\hat{I}'}_{t}$. Note that both the encoders and the decoder has a topology similar to that described in \secref{sec:coarse_inpainting}. Similar to the coarse inpainting network, we use the $L_1$ pixel-wise reconstruction loss $\mathcal{L}^{\bm{\hat{I}}}_{L_1}$ to supervise this sub-network. However, since the task here is image-to-image translation, we compute the loss over all the pixels in the image. To focus on the consistency of image patch features, we also use the perceptual loss $\mathcal{L}^{\bm{\hat{I}}}_{\Psi}$ and style loss $\mathcal{L}^{\bm{\hat{I}}}_{\text{G}}$ computed as\looseness=-1
\begin{align}
    \mathcal{L}^{\bm{\hat{I}}}_{\Psi} &= \sum_{ i } \left\lVert \bm{\hat{\Psi}}_{i}^{\scriptscriptstyle \text{VGG}} - \bm{\Psi}_{i}^{\scriptscriptstyle \text{VGG}} \right\rVert_1,\\
    \mathcal{L}^{\bm{\hat{I}}}_{\text{G}} &= \sum_{ i } \left\lVert \bm{\hat{G}}_{i}^{\scriptscriptstyle \text{VGG}} - \bm{G}_{i}^{\scriptscriptstyle \text{VGG}} \right\rVert_1,
\end{align}
where $\bm{\Psi}_{i}^{\scriptscriptstyle \text{VGG}}$ and $\bm{\hat{\Psi}}_{i}^{\scriptscriptstyle \text{VGG}}$ represent the VGG-16~\cite{simonyanZ14} feature maps for the groundtruth image without dynamic objects and the output of the refinement image-to-image translation network. Similarly, $\bm{G}_{i}^{\scriptscriptstyle \text{VGG}}$ and $ \bm{\hat{G}}_{i}^{\scriptscriptstyle \text{VGG}} $ are their corresponding Gram matrices at layer $i \in \{ \text{\textit{relu2\_2}}, \, \text{\textit{relu3\_3}},\, \text{\textit{relu4\_3}} \}$.

For adversarial training, we use a discriminator that takes the groundtruth inpainted image $\bm{I'}_{t}$, the inpainted output image $\bm{\hat{I}'}_{t}$ and the corresponding target region mask $\bm{M}_{t}$, concatenated channel-wise. The discriminator consisting of six sequential strided convolution layers with kernel size 5 and stride 2, is used to learn and discriminate feature statistics of Markovian patches producing a 3D feature map tensor $\bm{Y}$ as output. We adopt the SN-PatchGAN hinge loss~\cite{yu2018free} $\mathcal{L}^{\bm{\hat{I}}}_{\scriptscriptstyle \text{GAN}}$ to train our model under the generative adversarial framework as
\begin{equation} \label{gan_loss2}
   \mathcal{L}^{\bm{\hat{I}}}_{\scriptscriptstyle \text{GAN}} = \mathcal{L}^{\scriptscriptstyle \text{G}}_{\scriptscriptstyle \text{GAN}} + \mathcal{L}^{\scriptscriptstyle \text{D}}_{\scriptscriptstyle \text{GAN}},
\end{equation}
where
\begin{equation}
     \mathcal{L}^{\scriptscriptstyle \text{G}}_{\scriptscriptstyle \text{GAN}} = -\frac{1}{\scriptscriptstyle C \times H \times W} \sum_{i, j, k} \bm{\hat{Y}}\left(i, j, k\right),
\end{equation}
\begin{equation}
    \begin{split}
     \mathcal{L}^{\scriptscriptstyle \text{D}}_{\scriptscriptstyle \text{GAN}} = \frac{1}{\scriptscriptstyle C \times H \times W}\biggl\{ \sum_{i, j, k} \operatorname{ReLU}{\left[1 - \bm{Y}\left(i, j, k\right)\right]} \\
     + \sum_{i, j, k} \operatorname{ReLU}{\left[1 + \bm{\hat{Y}}\left(i, j, k\right)\right]} \biggr\},
     \end{split}
\end{equation}
$\bm{\hat{Y}} \in \mathbb{R}^{C \times H \times W}$ and $\bm{Y} \in \mathbb{R}^{C \times H \times W}$ are outputs of the image discriminator for both the fake input $\bm{\hat{I}'}_{\,t}$ and the real input $\bm{I'}_{\,t}$. We employ the SN-PatchGAN hinge loss element-wise on the output feature map, effectively defining $C \times H \times W$ discriminators focusing on different locations and different semantics. The receptive field of each neuron in the output map covers the entire input along its spatial dimensions, therefore separate global and local discriminators are not required.

The overall loss function of the refinement network is computed as
\begin{equation}
     \mathcal{L}^{\bm{\hat{I}}} =
     \lambda^{\bm{\hat{I}}}_{L_1} \mathcal{L}^{\bm{\hat{I}}}_{L_1} +
     \lambda^{\bm{\hat{I}}}_{\Psi} \mathcal{L}^{\bm{\hat{I}}}_{\Psi} +
     \lambda^{\bm{\hat{I}}}_{G} \mathcal{L}^{\bm{\hat{I}}}_{G} + 
     \lambda^{\bm{\hat{I}}}_{\scriptscriptstyle \text{GAN}} \mathcal{L}^{\bm{\hat{I}}}_{\scriptscriptstyle \text{GAN}},
\end{equation}
where $\lambda^{\bm{\hat{I}}}_{L_1}$, $\lambda^{\bm{\hat{I}}}_{\Psi}$, $\lambda^{\bm{\hat{I}}}_{G}$
and $\lambda^{\bm{\hat{I}}}_{\scriptscriptstyle \text{GAN}}$ are the loss weighting factors.

\subsection{Depth Completion}
\label{sec:depth_completion}

In order to regress the depth values in regions occluded by dynamic objects, we build upon a sparse-to-dense depth completion network~\cite{ma2018self} and adapt it to the depth inpainting task. The sub-network is a self-supervised deep regression model that obtains the supervision signal for regressing the missing depth values through inverse image warping. The architecture follows an encoder-decoder topology with skip connections, where the encoder is based on the ResNet-18 model~\cite{DBLP:conf/eccv/HeZRS16} and takes the inpainted image $\hat{I'}_{t}$ concatenated with the corresponding depth map $D_t$ with pixels in regions containing dynamic object set to zero using the binary target region mask $M_t$ from \secref{sec:coarse_inpainting}. The decoder consists of $3\times 3$ transpose convolutions, each of which upsamples the feature maps by a factor of two while halving the number of channels and fusing the corresponding encoder feature maps through skip connections. Finally, a $1\times 1$ convolution reduces the number of feature channels to one and yields the inpainted depth map at the same resolution as the input.

The task of depth completion is in the core of our geometry-aware approach, with inpainted depth maps providing the network with information about the scene geometry. The geometry awareness of our model comes from the fact that inpainted RGB images and inpainted depth maps supervise each other, as explained in \secref{sec:refinement}. Consequently, inpainting the current RGB frame using the color information from previous frames is possible only if previous frames have been properly warped, \textit{i.e.} only if the depth completion is correct. While RGB supervises depth completion in a form of a network input, the inpainted depth map supervises the inpainting of subsequent RGB frame through forward warping.

We train the depth completion network by optimizing the pixel-wise reconstruction loss $\mathcal{L}^{\bm{D}}_{L_1}$ and a smoothness loss $\mathcal{L}^{\bm{D}}_{\text{smooth}}$ as
\begin{equation}
    \mathcal{L}^{\bm{D}}_{L_1} = \left\lVert \left(\bm{\hat{D'}}_t - \bm{D'}_t\right) \odot \bm{M}_t \right\rVert_1,
\end{equation}
\begin{equation}
    \mathcal{L}^{\bm{D}}_{\text{smooth}} = \left\lVert \nabla^2 \bm{\hat{D'}}_t \right\rVert_1,
\end{equation}
where $\bm{D'}_t$ is the groundtruth depth map without dynamic objects and $\bm{\hat{D'}}_t$ is the output of the depth completion network. The reconstruction loss computes the $L_1$ distance between the prediction $\bm{\hat{D'}}_t$ and the groundtruth $\bm{D'}_t$ for pixels in the target regions. While the smoothness loss penalizes the $L_1$-norm of the Laplacian of predicted depth map $\bm{\hat{D'}}_t$ to encourage smooth predictions, i.e. to avoid discontinuities and introduce neighboring constraints.

The overall loss function of the depth completion network is computed as
\begin{equation}
    \mathcal{L}^{\bm{D}} = \lambda^{\bm{D}}_{L_1} \mathcal{L}^{\bm{D}}_{L_1} + \lambda^{\bm{D}}_{\text{smooth}} \mathcal{L}^{\bm{D}}_{\text{smooth}},
\end{equation}
where $\lambda^{\bm{D}}_{L_1}$ and $\lambda^{\bm{D}}_{\text{smooth}}$ are the loss weighting factors.

\subsection{Loss-Aware Scheduled Teacher Forcing}
\label{tf}

\begin{figure*}
\centering
\footnotesize
\setlength{\tabcolsep}{0.05cm}
{\renewcommand{\arraystretch}{0.3}
\begin{tabular}{P{0.5cm}P{2.75cm}P{2.75cm}P{2.75cm}P{2.75cm}P{2.75cm}P{2.75cm}}
& \raisebox{-0.4\height}{RGB (D)} & \raisebox{-0.4\height}{Depth (D)} & \raisebox{-0.4\height}{Semantic Mask (D)} & \raisebox{-0.4\height}{RGB (S)} & \raisebox{-0.4\height}{Depth (S)} & \raisebox{-0.4\height}{Semantic Mask (S)} \\
\\
\\
(a) & \raisebox{-0.4\height}{\includegraphics[width=\linewidth]{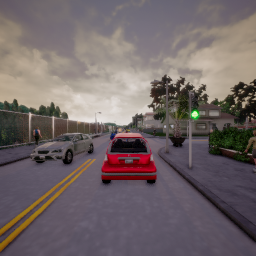}} & 
\raisebox{-0.4\height}{\includegraphics[width=\linewidth]{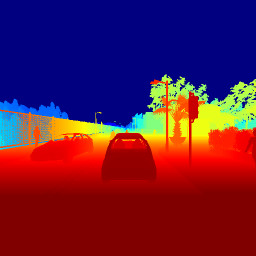}} & 
\raisebox{-0.4\height}{\includegraphics[width=\linewidth]{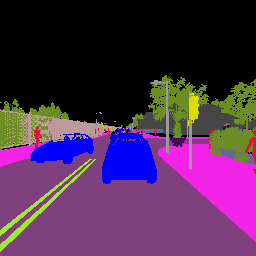}} & 
\raisebox{-0.4\height}{\includegraphics[width=\linewidth]{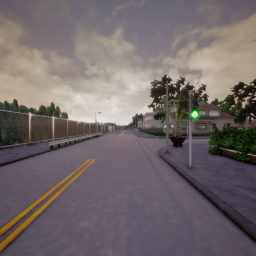}} & 
\raisebox{-0.4\height}{\includegraphics[width=\linewidth]{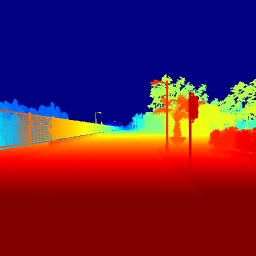}} & \raisebox{-0.4\height}{\includegraphics[width=\linewidth]{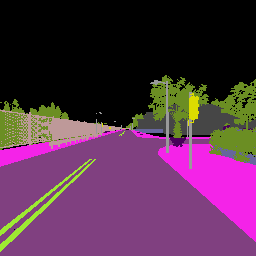}} \\
\\
\\
(b) & \raisebox{-0.4\height}{\includegraphics[width=\linewidth]{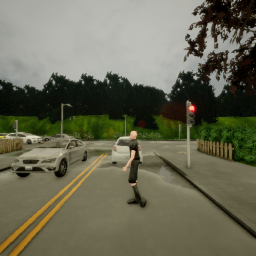}} & 
\raisebox{-0.4\height}{\includegraphics[width=\linewidth]{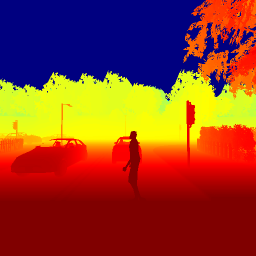}} & 
\raisebox{-0.4\height}{\includegraphics[width=\linewidth]{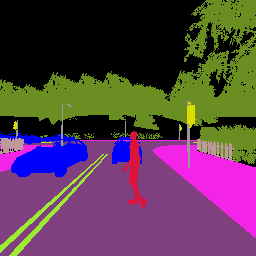}} & 
\raisebox{-0.4\height}{\includegraphics[width=\linewidth]{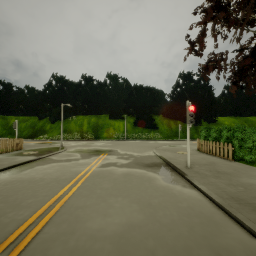}} & 
\raisebox{-0.4\height}{\includegraphics[width=\linewidth]{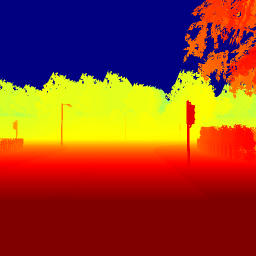}} & \raisebox{-0.4\height}{\includegraphics[width=\linewidth]{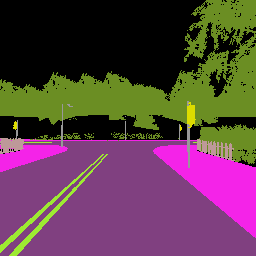}} \\
\\
\\
(c) &
\raisebox{-0.4\height}{\includegraphics[width=\linewidth]{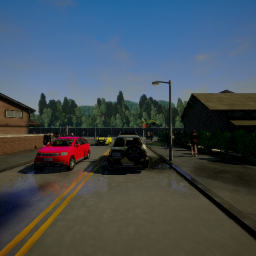}} & 
\raisebox{-0.4\height}{\includegraphics[width=\linewidth]{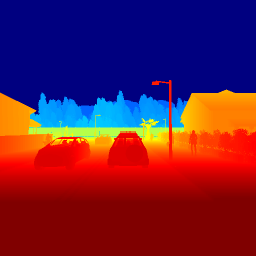}} & 
\raisebox{-0.4\height}{\includegraphics[width=\linewidth]{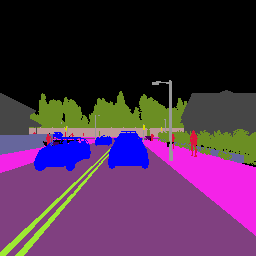}} & 
\raisebox{-0.4\height}{\includegraphics[width=\linewidth]{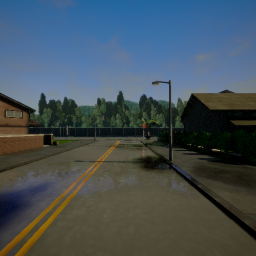}} & 
\raisebox{-0.4\height}{\includegraphics[width=\linewidth]{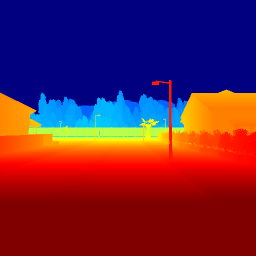}} & \raisebox{-0.4\height}{\includegraphics[width=\linewidth]{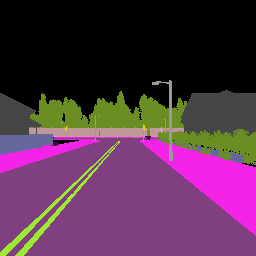}} \\
\end{tabular}}
\caption{Examples from our dataset showing paired dynamic (D) and static (S) scenes. Each image has a corresponding depth map, semantic segmentation labels and camera pose information. The first three columns show frames containing dynamic objects while the next three columns show their perfectly aligned static counterpart. Rows (a) - (c) visualize the diversity of the dataset across different weather conditions.}
\label{fig:dataset}
\end{figure*}

Given the fact that our overall architecture functions in a recurrent manner, we train our model with a variant of teacher forcing that emerges from the maximum likelihood criterion. Explicitly scheduling the decay probability of teacher forcing requires a prior estimate about the speed of convergence of the optimization process. However, providing a good estimate while learning very complex tasks is nearly infeasible. Nevertheless, we can determine a good value of the loss function or a metric at which the model can be considered to be fully trained for a given dataset. We use this insight and introduce a simple extension called loss-aware scheduled teacher forcing. Instead of defining the decay schedule as a function of the number of iterations, we decay the teacher forcing probability based on the current value of the loss function. In our case, we use the mean of $\mathcal{L}^{\bm{\hat{I}}}_{L_1}$ of the last 20 mini-batches to linearly decay the probability of teacher forcing $p_{\scriptscriptstyle TF}$ as
\begin{equation}
    p_{\scriptscriptstyle TF} = \begin{cases}
    1, & \mathcal{L}^{\bm{\hat{I}}}_{L_1} > d_{\text{start}} \\
    \frac{\mathcal{L}^{\bm{\hat{I}}}_{L_1} -  d_{\text{end}}}{d_{\text{start}} - d_{\text{end}}}, & d_{\text{start}} \geq \mathcal{L}^{\bm{\hat{I}}}_{L_1} \geq d_{\text{end}} \\
    0, & \mathcal{L}^{\bm{\hat{I}}}_{L_1} < d_{\text{end}}
    \end{cases},
\end{equation}
where $d_{\text{start}}$ and $d_{\text{end}}$ denote loss values at which the decay starts and ends. Note that here we consider loss functions (and metrics) that are monotonically decreasing over time.

\section{Experimental Evaluation}
\label{sec:experiments}

In this section, we first describe the data collection methodology in \secref{sec:dataset} and the training procedure that we employ in \secref{sec:training}. We then present quantitative comparisons of our DynaFill model against state-of-the-art methods in \secref{sec:comparisonSOTA}, followed by detailed ablation studies to gain insight on the improvement in performance due to various architectural components in \secref{sec:ablation}. Subsequently, we present visualizations of the learned gating mask in the recurrent gated feedback mechanism in \secref{sec:appendixGatingViz} and qualitative results in \secref{sec:qualitative}. Finally, we present a case study on employing our model as a preprocessor for retrieval-based visual localization in dynamic urban environments in \secref{sec:localization}.

\subsection{Dataset}
\label{sec:dataset}

As there are no publicly available RGB-D datasets with groundtruth for inpainting dynamic objects in urban scenes, we generated a hyperrealistic synthetic dataset using the CARLA~\cite{Dosovitskiy17} simulator. In order to have pixel perfect aligned frames, we run two instances of the CARLA simulator in parallel. The first instance with a maximum number of dynamic actors and the second instance without any dynamic actors. The instances were run in a synchronous mode where they wait for a control input before simulating the next frame. To ensure both the instances are synchronized, we used autopilot controls provided by the first simulation in both of the instances. Since the simulations tend to diverge over time, we keep track of the error between the poses of the two recording vehicles and automatically restarted the instances if the error threshold exceeds. We set the threshold empirically to $\SI{0.005}{\meter}$ (in $L_1$ space) and $\SI{0.01}{\degree}$. In order to uniformly cover the entire map, regardless of the random decisions of the autopilot, we manually modelled the map as a graph with each intersection being a node and each street being an edge. This enables us to terminate the data generation process after all streets (i.e. edges) have been visited.

Our dataset consists of 6-DoF groundtruth poses and aligned RGB-D images with and without dynamic objects, as well as groundtruth semantic segmentation labels. The dataset was collected in several weather conditions including ClearNoon, CloudyNoon, WetNoon, WetCloudyNoon, ClearSunset, CloudySunset, WetSunset, and WetCloudySunset. \figref{fig:dataset} shows examples from our dataset in the different weather conditions. The images were acquired at a resolution of $512\times 512$ pixels with a field of view of $\SI{90}{\degree}$ using a front-facing camera mounted on the car. The images are formatted as 3 channels with integer values in range of $[0, 255]$ and the depth maps have real number values normalized to the range of $[0, 1]$ where $1$ represents the maximum measureable distance determined by CARLA which is $\SI{1000}{\meter}$. The semantic segmentation labels are formatted as single channel images with integer values in range of $[0, 12]$ which represents 13 semantic classes. The semantic classes contained in the dataset are \textit{building}, \textit{fence}, \textit{pole}, \textit{road line}, \textit{road}, \textit{sidewalk}, \textit{vegetation}, \textit{wall}, \textit{traffic sign}, \textit{other}, \textit{pedestrian}, \textit{vehicle} and an \textit{ignore} class. Moreover, we provide 6-DoF camera poses for each of the frames in the dataset which are represented as 6-dimensional vectors containing $x, y, z$ coordinates expressed in meters and roll, pitch and yaw angles expressed in degrees ranging from $\SI{-180}{\degree}$ to $\SI{180}{\degree}$. The camera poses are defined in the coordinate system of Unreal Engine which is left-handed with $x$ being forward, $y$ right and $z$ up. The image data was collected using a front-facing camera mounted on the vehicle with a relative translation $\bm{t} = \irow{2.0 & 0.0 & 1.8}\SI{}{\meter}$ in the local vehicle coordinate frame. We obtain the camera intrinsics from CARLA's Unreal Engine parameters. The images were acquired at $\SI{10}{\hertz}$ and we split the data into training and validation sets. The training set was collected in the \textit{Town01} map and consists of $77,742$ RGB-D images. While the validation set was collected in the \textit{Town02} map and consists of $23,722$ RGB-D images. The dataset that we introduce in this work is the first temporal RGB-D inpainting dataset which consists of aligned dynamic scenes and their static equivalent. We make the dataset, code and models publicly available at \url{http://rl.uni-freiburg.de/research/rgbd-inpainting}.

\subsection{Training Protocol}
\label{sec:training}

\begin{table*}
\footnotesize
\centering
\caption{Performance comparison of our DynaFill model against image as well as video inpainting methods.}
\label{tab:baselines}
\begin{tabular}{p{5cm}|P{1cm}P{1cm}P{1cm}P{1cm}P{1cm}P{1cm}|P{1.5cm}|P{1.5cm}}
\toprule
Method & L1 $\downarrow$ & FID $\downarrow$ & PSNR $\uparrow$ & SSIM $\uparrow$ & FVD $\downarrow$ & LPIPS $\downarrow$ & Time [\si{\milli\second}] $\downarrow$ & RMSE [\SI{}{\meter}] $\downarrow$ \\
\noalign{\smallskip}\hline\hline\noalign{\smallskip}
Empty Cities~\cite{bescos2019empty} & 0.0058 & 3.6940 & 37.8351 & 0.9714 & 200.5015 & 0.0248 & \textbf{12.3038} & -\\
Context Encoders~\cite{pathakCVPR16context} & 0.0068 & 5.6306 & 34.4159 & 0.9590 & 214.0140 & 0.0272 & 23.3914 & -\\
DeepFill~v2~\cite{yu2018free} + BTS~\cite{lee2019big} & 0.0080 & 3.3183 & 36.1598 & 0.9647 & 178.2610 & 0.0345 & 24.4074 & 9.8063 \\
\midrule
Deep Video Inpainting~\cite{kim2019deep} & 0.0452	& 16.0718 & 28.5660 & 0.7643 & 346.0854 & 0.2207 & 85.2701 & -\\
Deep Flow-Guided Video Inpainting~\cite{Xu_2019_CVPR} & 0.0143 & 17.3782 & 33.0311 & 0.9632 & 903.7222 & 0.0343 & 1597.5020 & -\\
LGTSM~\cite{chang2019learnable} + BTS~\cite{lee2019big} & 0.0075 & 3.5670 & 36.5204 & 0.9701 & 344.8208 & 0.0231 & 226.0164 & 9.9751 \\
\midrule
\textbf{DynaFill (Ours)} & \textbf{0.0051} & \textbf{1.8571} & \textbf{39.5513} & \textbf{0.9780} & \textbf{143.6950} & \textbf{0.0172} & 52.1756 & \textbf{7.7820} \\
\bottomrule
\end{tabular}
\end{table*}

We train our model on RGB-D images of $256\times 256$ pixels resolution with groundtruth odometry and object masks. We employ a series of data augmentations on the RGB images, with parameters sampled uniformly within specific ranges. We modulate the brightness $[0.7. 1.3]$, contrast $[0.8, 1.2]$, saturation $[0.8, 1.2]$ and hue $[-0.15, 0.15]$. Additionally, we also randomly horizontally flip the images. We use the groundtruth odometry in the recurrent gated feedback while training and estimates from \cite{radwan2018vlocnetpp} during inference. We use the PyTorch deep learning library for implementing our DynaFill architecture and we train the model on NVIDIA TITAN X GPUs.

The overall loss function $\mathcal{L}$ that we use to optimize the DynaFill architecture can be expressed as
\begin{multline*}
     \mathcal{L} =
     \mathcal{L}^{\bm{\widetilde{I}}} + 
     \lambda^{\bm{\hat{I}}}_{L_1} \mathcal{L}^{\bm{\hat{I}}}_{L_1} +
     \lambda^{\bm{\hat{I}}}_{\Psi} \mathcal{L}^{\bm{\hat{I}}}_{\Psi} +
     \lambda^{\bm{\hat{I}}}_{G} \mathcal{L}^{\bm{\hat{I}}}_{G} + 
     \lambda^{\bm{\hat{I}}}_{\scriptscriptstyle \text{GAN}} \mathcal{L}^{\bm{\hat{I}}}_{\scriptscriptstyle \text{GAN}} + \\
     \lambda^{\bm{D}}_{L_1} \mathcal{L}^{\bm{D}}_{L_1} + \lambda^{\bm{D}}_{\text{smooth}} \mathcal{L}^{\bm{D}}_{\text{smooth}},
\end{multline*}
where the loss weighting factors for the refinement image-to-image translation sub-network are $\lambda^{\bm{\hat{I}}}_{L_1} = 1.0$, $\lambda^{\bm{\hat{I}}}_{\Psi} = 0.3$, $\lambda^{\bm{\hat{I}}}_{G} = 0.3$, and $\lambda^{\bm{\hat{I}}}_{\scriptscriptstyle \text{GAN}} = 1.0$. Analogously, the loss weighting factors for the depth completion sub-network are $\lambda^{\bm{D}}_{L_1} = 0.01$ and $\lambda^{\bm{D}}_{\text{smooth}} = 0.001$.

During the training, we use our loss-aware extension of scheduled teacher forcing. Instead of defining the decay schedule as a function of the total number of iterations, we decay the teacher forcing probability based on the value of the loss function, for which it is much easier to estimate the final value. We set $d_{\text{start}} = 0.06$ and $d_{\text{end}} = 0.01$. We update the weights of the network with the ADAM optimizer with an initial learning rate of $\alpha = {10}^{-4}$, and first and second momentum decay rates of $\beta_1=0.5$ and $\beta_2=0.999$ respectively. In order to ease the optimization, we first pre-train each of the sub-networks: depth completion with a mini-batch size of 24 on two GPUs, coarse inpainting with a mini-batch size of 104 on four GPUs and refinement image-to-image translation with a mini-batch size of 48 on four GPUs. We then fine-tune the entire architecture by initializing the model with the aforementioned pre-trained weights with a mini-batch size of 36 on four GPUs. Additionally, we use early stopping with a patience of 10 epochs to avoid overfitting.

\subsection{Comparison with the State-of-the-Art}
\label{sec:comparisonSOTA}

\begin{table}
\footnotesize
\centering
\caption{Comparison with image-to-image synthesis methods.}
\label{tab:baselines_i2i}
\begin{tabular}{p{2cm}|P{0.8cm}P{0.8cm}P{0.7cm}P{0.8cm}P{0.8cm}P{0.8cm}}
\toprule
Method & FID & PSNR & SSIM & FVD & LPIPS \\
\noalign{\smallskip}\hline\hline\noalign{\smallskip}
Empty Cities~\cite{bescos2019empty} & 2.4112 & 37.2937 & 0.9479 & 85.7118 & 0.0376 \\
Context Enc.~\cite{pathakCVPR16context} & 113.5832 & 21.9931 & 0.4637 & 1724.43 & 0.4359 \\
DeepFill~v2~\cite{yu2018free} & 1.0836 & 36.6502 & 0.9464 & 85.2216 & 0.0270 \\
\midrule
\textbf{DynaFill (Ours)} & \textbf{1.0025} & \textbf{38.8672} & \textbf{0.9651} & \textbf{27.8344} & \textbf{0.0124} \\
\bottomrule
\end{tabular}
\end{table}

\begin{table*}
\footnotesize
\centering
\begin{threeparttable}
\caption{Ablation study on the topology of our DynaFill architecture showing the impact due to the various network components.}
\label{tab:ablation_i2i}
\begin{tabular}{P{0.7cm}|P{0.5cm}P{0.5cm}P{0.5cm}P{0.5cm}P{0.5cm}|P{1cm}P{1cm}P{1cm}P{1cm}P{1cm}P{1cm}|P{1.5cm}}
\toprule
Model & \multicolumn{5}{c}{Configuration} & \multicolumn{6}{|c|}{RGB} & Depth \\
 \cmidrule{2-13}
 & JT & TF & RN & GAN & PSL & L1 $\downarrow$ & FID $\downarrow$ & PSNR $\uparrow$ & SSIM $\uparrow$ & FVD $\downarrow$ & LPIPS $\downarrow$ & RMSE [\SI{}{\meter}] $\downarrow$ \\
\noalign{\smallskip}\hline\hline\noalign{\smallskip}
A & - & - & - & - & - & 0.0270 & 9.4045 & 33.4402 & 0.9175 & 120.5195 & 0.0587 & 9.3649 \\
B & \checkmark & - & - & - & - & 0.0172 & 6.0380 & 35.8415 & 0.9374 & 128.2434 & 0.0334 & 8.9366 \\
C & \checkmark & \checkmark & - & - & - & 0.0170 & 4.6829 & 36.1967 & 0.9425 & 98.5418 & 0.0272 & 9.3043 \\
D & \checkmark & \checkmark & \checkmark & - & - & 0.0158 & 3.7079 & 36.5717 & 0.9441 & 53.2797 & 0.0252 & 8.1100 \\
E & \checkmark & \checkmark & \checkmark & \checkmark & - & 0.0123 & 1.4747 & 38.2600 & 0.9600 & 46.9936 & 0.0164 & 7.8952 \\
F & \checkmark & \checkmark & \checkmark & \checkmark & \checkmark & \textbf{0.0112} & \textbf{1.0025} & \textbf{38.8672} & \textbf{0.9651} & \textbf{27.8344} & \textbf{0.0124} & \textbf{7.7701} \\
\bottomrule
\end{tabular}
JT = Coarse inpainting and depth completion trained jointly, TF = Teacher forcing, RN = Refinement network, PSL = Perceptual \& Style losses
\end{threeparttable}
\end{table*}

As there are no end-to-end learning-based RGB-D video inpainting techniques, we compare against both single image inpainting methods (Empty Cities~\cite{bescos2019empty}, Context Encoders~\cite{pathakCVPR16context}, DeepFill~v2~\cite{yu2018free}) as well as video inpainting methods (Deep Video Inpainting~\cite{kim2019deep}, Deep Flow-Guided Video Inpainting~\cite{Xu_2019_CVPR}, LGTSM~\cite{chang2019learnable}).
Additionally, we create two strong RGB-D inpainting baselines to compare against, by combining the state-of-the-art image inpainting model DeepFill~v2~\cite{yu2018free} with the state-of-the-art monocular depth prediction approach BTS~\cite{lee2019big}, and the state of the art video inpainting approach LGTSM~\cite{chang2019learnable} with BTS. Here the inputs to the depth prediction networks were the inpainted images from DeepFill~v2 or LGTSM, respectively. We used the publicly available implementations of these networks to train the models on our proposed dataset. We report the quantitative performance in terms of several standard metrics that cover both image and video inpainting quality, namely L1 distance, Fr\'{e}chet Inception Distance (FID)~\cite{heusel2017gans}, Peak Signal-to-Noise Ratio (PSNR), Structural Similarity Index Measure (SSIM)~\cite{wang2004image}, Fr\'{e}chet Video Distance (FVD)~\cite{unterthiner2018towards} and Learned Perceptual Image Patch Similarity (LPIPS)~\cite{zhang2018unreasonable}. We present results for depth inpainting in terms of the Root-Mean-Square Error (RMSE) in \tabref{tab:baselines} and \tabref{tab:ablation_i2i}.\looseness=-1

As our model performs both inpainting and image-to-image translation, we perform two sets of evaluations by computing the metrics only for the inpainted region (inpainting task) and for the entire image (image-to-image translation task). \tabref{tab:baselines} and \tabref{tab:baselines_i2i} shows comparisons from this experiment. We observe that our proposed DynaFill model exceeds the performance of competing methods in all the metrics and for both tasks, thereby achieving state-of-the-art performance. For inpainting, our model achieves an improvement of $1.46$ in the FID score and $34.57$ in the FVD score over the previous state-of-the-art DeepFill~v2. While for image-to-image translation, DynaFill achieves an improvement of $0.08$ in the FID score and $57.39$ in the FVD score. The large improvement in the FVD scores demonstrate the temporal consistency achieved by our method, while still being faster in inference than other video inpainting methods that even use future frame information.
Additionally, the lower value of the RMSE metric achieved by our DynaFill model in comparison to the RGB-D baselines, indicates that our approach effectively utilizes depth information, as opposed to just performing direct depth completion on top of the inpainted RGB image. Our model improves the depth inpainting by more than $\SI{2}{\meter}$ in RMSE over (DeepFill~v2 + BTS) and (LGTSM + BTS).

\subsection{Ablation Studies}
\label{sec:ablation}

In this section, we systematically study the impact of various architectural components in our proposed DynaFill model. We use FID and FVD scores as the primary evaluation metrics for temporal image inpainting and the RMSE metric for depth inpainting. However, we also report the other image and video inpainting metrics for completeness.

\subsubsection{Detailed Study on the DynaFill Architecture}

In \tabref{tab:ablation_i2i}, we present results of the ablation study to assess the influence of various design choices on the performance of our proposed image-to-image translation model. The basic model~A consisting of disjoint individually trained coarse inpainting and depth completion networks only with the pixel-wise $L_1$ reconstruction loss and depth smoothness loss achieves an FID score of $9.40$, FVD score of $120.52$ and a RMSE of $\SI{9.36}{\meter}$. In model~B, we concatenate the inpainted image with the depth map and feed it as an input to the depth completion network, and train it end-to-end with the coarse inpainting network. This leads to a small drop of $3.37$ in the FID score and an increase of $7.72$ in the FVD score which indicates that the smoothness loss employed on the depth completion network causes the coarse inpainting network to produce a blurry output. However, this improves the performance of the depth completion network yielding a RMSE of $\SI{8.94}{\meter}$. In order to optimize the joint model more effectively, we employ our loss-aware scheduled teacher forcing in model~C which also prevents the gradient from the depth completion network to flow into the coarse inpainting network. This leads to an improvement over model A in both the FID and FVD scores by $1.36$ and $29.70$ respectively.

Subsequently, we introduce temporal context in model~D by incorporating our recurrent gated feedback mechanism in the refinement network. This leads to an improvement of $0.97$ in the FID score, $45.26$ in the FVD score, and a large improvement yielding a RMSE of $\SI{8.11}{\meter}$. These resulting metrics show the importance of mutual supervision of RGB and depth networks, and employing inpainted depth maps to achieve the geometry-aware approach. Visualizations analyzing the learning gating masks are shown in \secref{sec:appendixGatingViz}. We then employ adversarial training in model~E which enforces the distributions of both spatial and temporal features of the generated frames to be indistinguishable from the groundtruth. This model achieves an FID score of $1.47$, a FVD score of $46.99$, and also reduces the RMSE by $\SI{0.21}{\meter}$ as the depth inpainting is conditioned on better image inpainting. Finally, we guide the optimization process by enforcing perceptual and style consistency of learned features which further improves the FID and FVD scores by $0.47$ and $19.16$ respectively, in addition to improving the RMSE by $\SI{0.13}{\meter}$.

\begin{table*}
\footnotesize
\centering
\begin{threeparttable}
\caption{Ablation study on the topology of DynaFill evaluated only for inpainting, showing the impact due to the various network components.}
\label{tab:ablation_inpainting}
\begin{tabular}{P{1cm}|P{0.7cm}P{0.7cm}P{0.7cm}P{0.7cm}P{0.7cm}|P{1.2cm}P{1.2cm}P{1.2cm}P{1.2cm}P{1.2cm}P{1.2cm}}
\toprule
Model & \multicolumn{5}{c}{Configuration} & \multicolumn{6}{|c}{RGB} \\
\cmidrule{2-12}
 & JT & TF & RN & GAN & PSL & L1 $\downarrow$ & FID $\downarrow$ & PSNR $\uparrow$ & SSIM $\uparrow$ & FVD $\downarrow$ & LPIPS $\downarrow$ \\
\noalign{\smallskip}\hline\hline\noalign{\smallskip}
A & - & - & - & - & - & 0.0059 & 4.0209 & 38.3690 & 0.9734 & 263.4455 & 0.0229 \\
B & \checkmark & - & - & - & - & 0.0061 & 5.6052 & 37.7146 & 0.9722 & 327.3409 & 0.0285 \\
C & \checkmark & \checkmark & - & - & - & 0.0055 & 3.4841 & 38.9247 & 0.9749 & 214.0470 & 0.0222 \\
D & \checkmark & \checkmark & \checkmark & - & - & 0.0051 & 2.1856 & 39.6628 & 0.9786 & 163.9389 & 0.0187 \\
E & \checkmark & \checkmark & \checkmark & \checkmark & - & 0.0051 & 2.1354 & 39.7122 & 0.9787 & 158.9624 & 0.0185 \\
F & \checkmark & \checkmark & \checkmark & \checkmark & \checkmark & \textbf{0.0051} & \textbf{1.8571} & \textbf{39.5513} & \textbf{0.9780} & \textbf{143.6950} & \textbf{0.0172} \\
\bottomrule
\end{tabular}
JT = Coarse inpainting and depth completion trained jointly, TF = Teacher forcing, RN = Refinement network, PSL = Perceptual \& Style losses
\end{threeparttable}
\end{table*}

\tabref{tab:ablation_inpainting} shows the ablation study for assessing how the design choices influence the model evaluated only for inpainting. Note that the performance of the depth completion network remains the same as the RMSE values reported in \tabref{tab:ablation_i2i}. The baseline model~A consisting of disjoint individually trained coarse inpainting and depth completion networks with pixel-wise $L_1$ reconstruction and depth smoothness losses respectively, achieve an FID score of $4.02$, an FVD score of $263.45$ and RMSE of $\SI{9.36}{\meter}$. In model~B, we concatenate the inpainted image with the depth map and feed it as an input to the depth completion network, and train it end-to-end with the coarse inpainting network which leads to a small drop in the FID score by $1.58$ and $63.90$ in the FVD score, however it improves the performance of the depth completion network yielding an RMSE of $\SI{8.94}{\meter}$. Notably, this shows that the smoothness loss causes the coarse inpainting sub-network to output blurry images. In model~C, we employ our loss-aware scheduled teacher forcing which yields an improvement over model~A in both FID and FVD scores by $0.54$ and $49.40$. This can be attributed to the fact that it helps improve the optimization and it additionally stops the gradient from the depth completion network to flow into the coarse inpainting network which prevents blurry outputs. Furthermore, in comparison to model~C, model~D achieves an improvement of $1.30$ in the FID score and $50.11$ in the FVD score by refining the coarsely inpainted frame and incorporating temporal context using our recurrent gated feedback mechanism. Subsequently, Model~E which is optimized in the generative adversarial framework achieves an FID score of $2.14$ and an FVD score of $158.96$. Finally, we enforce perceptual and style consistency in model~F which further leads to an improvement of $0.28$ and $15.27$ in the FID and FVD scores respectively.

We observe that the overall performance of the models in this ablation study is worse while evaluating for inpainting in comparison to image-to-image translation which is reported in \tabref{tab:ablation_i2i}. This can be attributed to the fact that inpainting models generate content only inside target regions while keeping the rest of the regions in the image intact. Therefore, they are unable to correct regions corelated with the presence of dynamic objects, namely shadows or reflections. Image-to-image translation models do correct regions surrounding the dynamic objects. Most importantly, this indicates that the evaluation metrics are able to distinguish between visually appealing and unappealing results.

\subsubsection{Evaluation with Varying Amounts of Noise}
\label{sec:noisy_data}

\usepgfplotslibrary{groupplots}
\begin{figure*}
    \centering
    \small
    \begin{tikzpicture}
    \begin{groupplot}[group style={group size=2 by 1, horizontal sep=0.9cm, vertical sep=0cm}]
    \nextgroupplot[%
        title={FID},
        xlabel={$p_{\text{n}}$},
        ylabel near ticks,
        xmin=0, xmax=1.0,
        xtick={0.0,0.1,0.2,0.3,0.4,0.5,0.6,0.7,0.8,0.9,1.0},
        ytick={1,2,3,4,5,6},
        legend pos=north west,
        legend cell align=left,
        legend style={
            nodes={scale=0.65, transform shape},
            draw=none,
            text opacity=1,
            fill opacity=0.5
        },
        smooth,
        ymajorgrids=true,
        xmajorgrids=true,
        grid style=dashed,
        height=4cm,
        width=7cm,
        scale only axis,
        tick label style={font=\footnotesize}
    ]
    \addplot[
        color=violet,
        mark=triangle*,
        ]
        coordinates {
            (0.0, 1.0025)
            (0.1, 1.0442)
            (0.2, 1.0634)
            (0.3, 1.1072)
            (0.4, 1.1747)
            (0.5, 1.2707)
            (0.6, 1.3777)
            (0.7, 1.5059)
            (0.8, 1.6559)
            (0.9, 1.8205)
            (1.0, 2.0041)
        };
    \addplot[
        color=red,
        mark=square*,
        ]
        coordinates {
            (0.0, 1.0025)
            (0.1, 1.7903)
            (0.2, 1.8922)
            (0.3, 2.1037)
            (0.4, 2.6037)
            (0.5, 3.2644)
            (0.6, 3.9353)
            (0.7, 4.5247)
            (0.8, 5.0338)
            (0.9, 5.4482)
            (1.0, 5.7283)
        };
    \addplot[
        color=blue,
        mark=*,
        ]
        coordinates {
            (0.0, 1.0025)
            (0.1, 1.0175)
            (0.2, 1.0291)
            (0.3, 1.0358)
            (0.4, 1.0481)
            (0.5, 1.0685)
            (0.6, 1.1222)
            (0.7, 1.1413)
            (0.8, 1.1699)
            (0.9, 1.2107)
            (1.0, 1.2471)
        };
    \addplot[color=red, dashed, thick] coordinates {(0,2.5877) (1,2.5877)};
    \addplot[color=red, dotted, thick] coordinates {(0,4.0001) (1,4.0001)};
    \legend{\textbf{Semantic mask}, \textbf{Odometry}, \textbf{Depth}, \textbf{SVO 2.0 (\textit{best})}, \textbf{SVO 2.0 (\textit{dynamic})}}
    \nextgroupplot[%
        title={FVD},
        xlabel={$p_{\text{n}}$},
        ylabel near ticks,
        xmin=0, xmax=1.0,
        xtick={0.0,0.1,0.2,0.3,0.4,0.5,0.6,0.7,0.8,0.9,1.0},
        ytick={50,100,150,200,250,300,350},
        legend pos=north west,
        legend cell align=left,
        smooth,
        ymajorgrids=true,
        xmajorgrids=true,
        grid style=dashed,
        height=4cm,
        width=7cm,
        scale only axis,
        tick label style={font=\footnotesize}
    ]
    \addplot[
        color=violet,
        mark=triangle*,
        ]
        coordinates {
            (0.0, 27.8344)
            (0.1, 35.773)
            (0.2, 45.408)
            (0.3, 58.0007)
            (0.4, 76.6354)
            (0.5, 96.5011)
            (0.6, 119.6857)
            (0.7, 148.5383)
            (0.8, 174.3261)
            (0.9, 204.1443)
            (1.0, 233.8535)
        };
    \addplot[
        color=red,
        mark=square*,
        ]
        coordinates {
            (0.0, 27.8344)
            (0.1, 63.2719)
            (0.2, 85.8232)
            (0.3, 120.6872)
            (0.4, 163.008)
            (0.5, 207.4282)
            (0.6, 245.8346)
            (0.7, 273.5215)
            (0.8, 296.6614)
            (0.9, 312.1913)
            (1.0, 317.4398)
        };
    \addplot[
        color=blue,
        mark=*,
        ]
        coordinates {
            (0.0, 27.8344)
            (0.1, 28.6003)
            (0.2, 28.7247)
            (0.3, 28.7105)
            (0.4, 29.0344)
            (0.5, 29.9216)
            (0.6, 30.0885)
            (0.7, 30.8219)
            (0.8, 31.2089)
            (0.9, 31.2061)
            (1.0, 32.1388)
        };
    \addplot[color=red, dashed, thick] coordinates {(0,102.8080) (1,102.8080)};
    \addplot[color=red, dotted, thick] coordinates {(0,312.7822) (1,312.7822)};
    \end{groupplot}
    \end{tikzpicture}
\caption{Effect of noisy inputs on the performance of our DynaFill model. While relatively accurate odometry is essential for ensuring both spatial and temporal consistency, the semantic mask noise causes instability only in the temporal domain due to mask flickering. Noisy depth maps, on the other hand, have negligible influence on the overall performance.}
\label{fig:noisy_data}
\end{figure*}
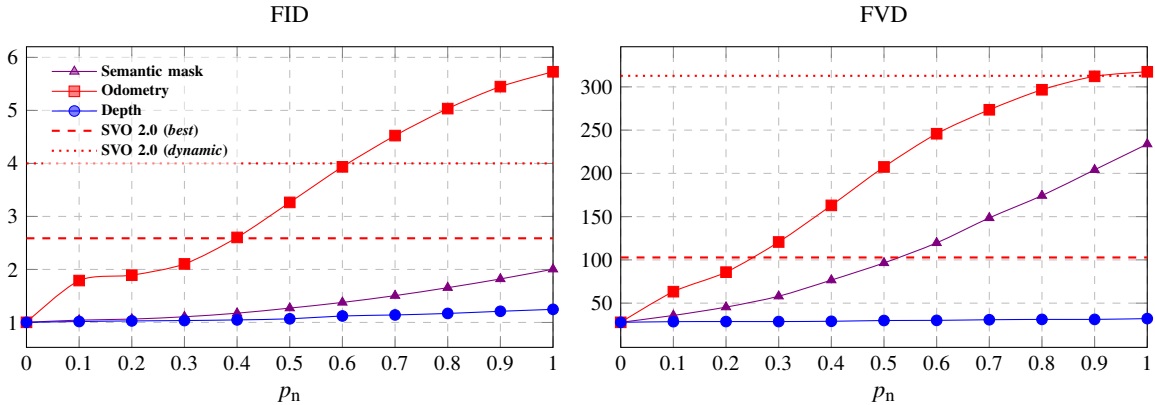

Measurements and estimates made from them in real-world scenarios are far from ideal, typically containing a certain amount of noise. In this section, we investigate the effects of noisy data inputs, namely semantic mask, odometry and depth, and report the performance by employing a real-world visual odometry system during inference.

{\parskip=5pt
\noindent\textbf{Noisy Inputs}: In order to compare effects of noise for each of the inputs, we introduce a single independent variable $p_{\text{n}} \in \left[0, 1\right]$ which we use to control the amount of noise. To model the semantic segmentation noise, we approximate each blob in a dynamic objects binary mask with $20\%$ of all pixels in the corresponding contour. We then offset each pixel $ \bm{p}_{ij} = \irow{u_{ij} & v_{ij}} $ from the $i$-th contour by $ \left(\bm{p}_{ij} - \bm{c}_i\right)/\lVert\bm{p}_{ij} - \bm{c}_i \rVert_2 \cdot \varepsilon_i $ where $ \bm{c}_i $ is the center pixel of the $i$-th blob and $ \varepsilon_i \sim \mathscr{N}\left(0, \left(p_{\text{n}} \sigma_{i} \right)^2 \right) $. Here, we use $ \sigma_{i} = r_{i} / 5 $, where $ r_{i} = \max_{j, k}{\lVert\bm{p}_{ij} - \bm{p}_{ik}\rVert_2} / 2$ is the radius of $i$-th contour.}

For depth noise, we first deform shapes in exactly the same way as for semantic masks but for pixels of all semantic classes (except road and sidewalk). Additionally, we apply Sobel filter in $x$-direction and set all pixels in the original depth map that are above the threshold $ T_{\,\text{Sobel}}$ to 0. In our case, we set $ T_{\,\text{Sobel}} $ = 5. Pixel-wise noise is then simulated using Kinetic depth noise model, where we also multiply its standard deviation by $ p_{\text{n}} $ and make sure that the offset does not exceed $ \SI{5}{\meter} $. In the end, all the pixels with the maximum depth value are also set to 0 with probability $ p_{\text{n}} $.

\begin{figure}
\centering
\footnotesize
\setlength{\tabcolsep}{0.05cm}
{\renewcommand{\arraystretch}{0.15}
\begin{tabular}{P{0.3cm}P{2cm}P{2cm}P{2cm}P{2cm}}
& \raisebox{-0.4\height}{0\%} & \raisebox{-0.4\height}{25\%} & \raisebox{-0.4\height}{50\%} & \raisebox{-0.4\height}{75\%} \\
\\
\\
\raisebox{0.2\height}{\rotatebox[origin=c]{90}{Depth}} &
\raisebox{-0.4\height}{\includegraphics[width=\linewidth]{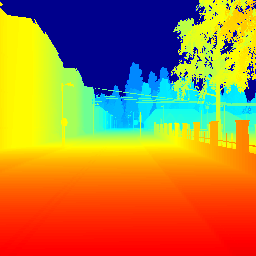}} & 
\raisebox{-0.4\height}{\includegraphics[width=\linewidth]{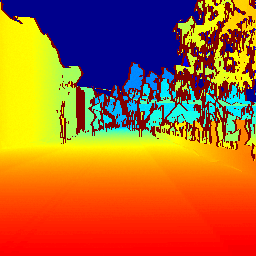}} & 
\raisebox{-0.4\height}{\includegraphics[width=\linewidth]{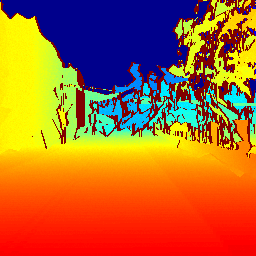}} & 
\raisebox{-0.4\height}{\includegraphics[width=\linewidth]{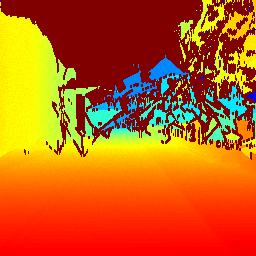}} \\
\\
\\
\raisebox{0.2\height}{\rotatebox[origin=c]{90}{Binary mask}} &
\raisebox{-0.4\height}{\includegraphics[width=\linewidth]{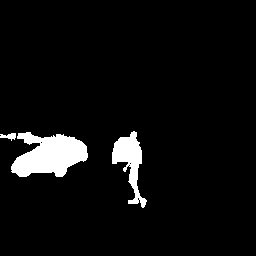}} & 
\raisebox{-0.4\height}{\includegraphics[width=\linewidth]{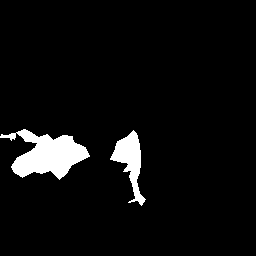}} & 
\raisebox{-0.4\height}{\includegraphics[width=\linewidth]{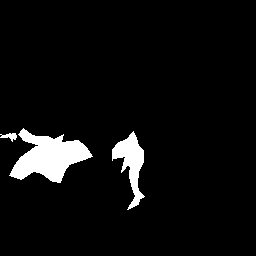}} & 
\raisebox{-0.4\height}{\includegraphics[width=\linewidth]{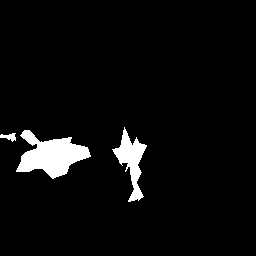}} \\
\end{tabular}}
\caption{Visualization of simulated noisy image inputs for various percentages of noise $ p_{\text{n}} $. We perform random shape deformations based on the semantic classes of objects by approximating them as 2D contours and adding offsets to the points they contain. Additionaly, for depth maps, we drop pixels near object boundaries and we employ pixel-wise noise on depth maps which is usually encoutered in real-world settings.}
\label{fig:noisy_inputs_visualization}
\end{figure}

Finally, we model odometry noise by adding $ \varepsilon_{\bm{t}} \sim \mathscr{N}\left(0, \left(p_{\text{n}} \sigma_{\bm{t}} \right)^2 \right) $ to positional ($x$, $y$, $z$) and $ \varepsilon_{\bm{R}} \sim \mathscr{N}\left(0, \left(p_{\text{n}} \sigma_{\bm{R}} \right)^2 \right)$ to rotational (roll, pitch, yaw) degrees of freedom. We set $ \sigma_{\bm{t}} = \SI{1}{\meter} $ and $ \sigma_{\bm{R}} = \SI{45}{\degree}$. To get a better understanding of artifacts that we induce in the inputs, we visualize the simulated noisy image data in \figref{fig:noisy_inputs_visualization}.

Results of experiments presented in \figref{fig:noisy_data} show the model is most sensitive to odometry noise. The results also show how crucial well-estimated odometry is in our approach which relies on properly aligned frames to be able to exploit the temporal context. However, both FID and FVD scores indicate that our system is robust enough to allow for $10\%$ to $20\%$ odometry noise margin to other competing methods. In contrast, the semantic mask noise is more tolerable, where the FID score show a diminished influence in the spatial domain which can be attributed to the refinement network being able to synthesize and correct regions that were not explicitly masked, as demonstrated in \secref{sec:qualitative}. However, the FVD scrores indicate a negative influence on the temporal consistency due to the semantic mask flickering between consecutive frames. Finally, the least sensitive input is the depth map for which both metrics demonstrate negligable decrease in performance (1.25 and 32.14 at $ p_{\text{n}}= 1 $ for FID and FVD scores, respectively).

{\parskip=5pt
\noindent\textbf{Odometry During Inference}: During inference time, the odometry has to be estimated from the incoming frames which are usually highly dynamic and the estimates often contains significant error. However, since our architecture predicts an inpainted RGB frame at each timestep, we can use this information to reduce the error induced by dynamic objects. For each timestep, we use best estimates of two consecutive inpainted RGB images. That is, we estimate the odometry between steps $t - 1$ and $t$ by using the refined inpainted RGB image from the previous timestep $\bm{\hat{I}'}_{t-1}$ and the coarsely inpainted RGB image from the current timestep $\bm{\widetilde{I}'}_{t}$. This is possible due to the odometry being required only in the refinement sub-network. For estimating the odometry, we use a monocular visual odometry system SVO 2.0~\cite{7782863}. We evaluate the performance of our DynaFill model for two cases: while using odometry predicted from the best inpainted images (\textit{best}) and while using odometry predicted from images containing dynamic objects (\textit{dynamic}). The corresponding metrics are used as reference values in \figref{fig:noisy_data}. This experiment indicates that using best inpainted images significantly improves the performance over estimating the odometry on raw dynamic frames, while the overall performance (in terms of FID and FVD) remains comparable with the closest baseline methods.}

\subsection{Visualization of Learned Gating Masks}
\label{sec:appendixGatingViz}

\begin{figure*}
\centering
\footnotesize
\setlength{\tabcolsep}{0.05cm}
{\renewcommand{\arraystretch}{0.3}
\begin{tabular}{P{0.5cm}P{2.75cm}P{2.75cm}P{2.75cm}P{2.75cm}P{2.75cm}P{2.75cm}}
& $\bm{I}_{t}$ & $\bm{D}_{t}$ & $\bm{\widetilde{I}'}_{t}$ & $\bm{M_{\Psi}}$ & $\bm{\hat{I}'}_{t}$ & $\bm{\hat{D'}}_t$ \\
\\
\\
(a) & \raisebox{-0.4\height}{\includegraphics[width=\linewidth]{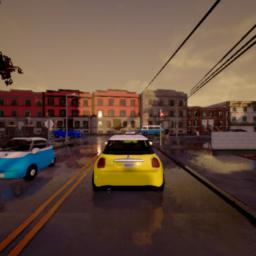}} & 
\raisebox{-0.4\height}{\includegraphics[width=\linewidth]{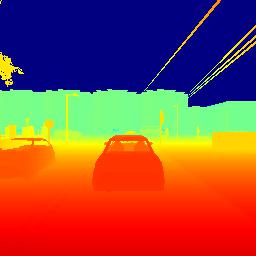}} &
\raisebox{-0.4\height}{\includegraphics[width=\linewidth]{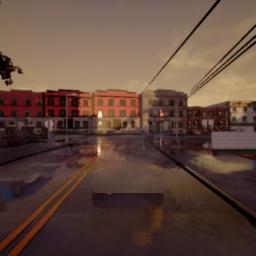}} &
\raisebox{-0.4\height}{\includegraphics[width=\linewidth]{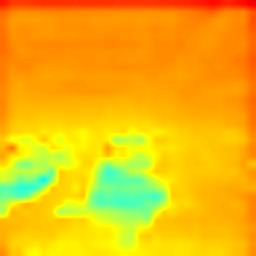}} &
\raisebox{-0.4\height}{\includegraphics[width=\linewidth]{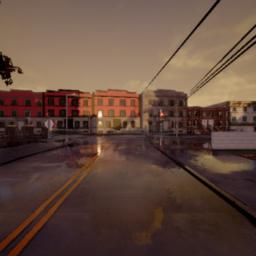}} &
\raisebox{-0.4\height}{\includegraphics[width=\linewidth]{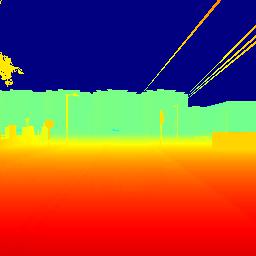}} \\
\\
\\
(b) &
\raisebox{-0.4\height}{\includegraphics[width=\linewidth]{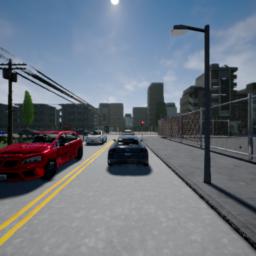}} & 
\raisebox{-0.4\height}{\includegraphics[width=\linewidth]{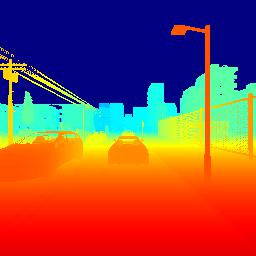}} & 
\raisebox{-0.4\height}{\includegraphics[width=\linewidth]{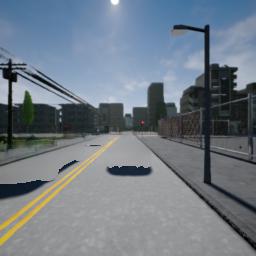}} &
\raisebox{-0.4\height}{\includegraphics[width=\linewidth]{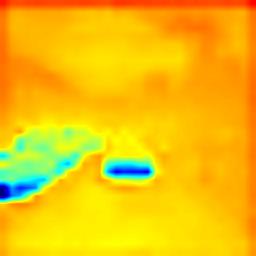}} &
\raisebox{-0.4\height}{\includegraphics[width=\linewidth]{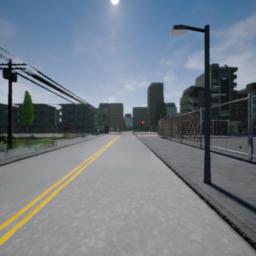}} &
\raisebox{-0.4\height}{\includegraphics[width=\linewidth]{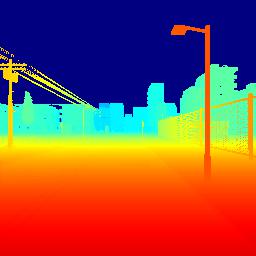}}\\
\\
\\
(c) &
\raisebox{-0.4\height}{\includegraphics[width=\linewidth]{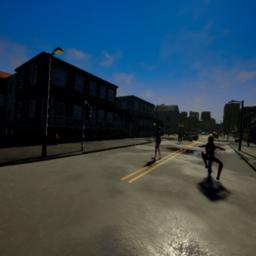}} & 
\raisebox{-0.4\height}{\includegraphics[width=\linewidth]{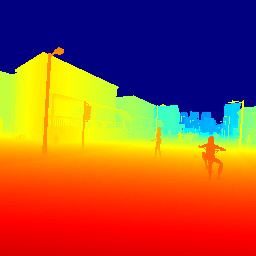}} & 
\raisebox{-0.4\height}{\includegraphics[width=\linewidth]{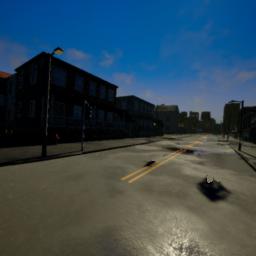}} &
\raisebox{-0.4\height}{\includegraphics[width=\linewidth]{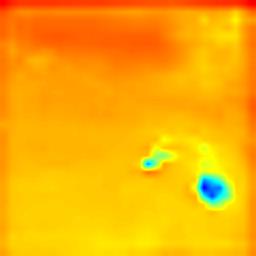}} &
\raisebox{-0.4\height}{\includegraphics[width=\linewidth]{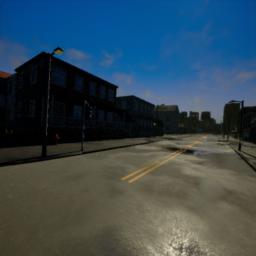}} &
\raisebox{-0.4\height}{\includegraphics[width=\linewidth]{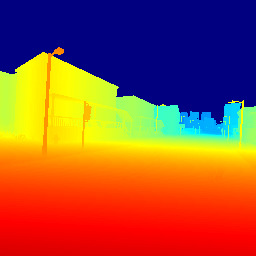}}\\
\\
\end{tabular}}
\caption{Visualization of the learned gating mask $\bm{M_{\Psi}}$ from the refinement image-to-image translation sub-network. We colorize the gating mask with the jet color map for better visibility. We also show the input image $\bm{I}_{t}$, the input depth map $\bm{D}_{t}$, the output of the coarse inpainting sub-network $\bm{\widetilde{I}'}_{t}$, the output of the refinement image-to-image translation sub-network $\bm{\hat{I}'}_{t}$, and the inpainted depth map $\bm{\hat{D'}}_t$. From these vizualizations, we can interpret the process that the network employs to fuse spatio-temporal features. We observe that gating masks indicate that the network primarily uses the previous frame information (indicated by blue) to correct regions that remain inconsistent after the removal of dynamic objects, namely shadows or reflections.}
\label{fig:gate}
\end{figure*}

Our refinement image-to-image translation sub-network employs our recurrent gated feedback mechanism that learns a gating mask which is utilized for fusing features from the inpainted image from the previous timestep and the coarsely inpainted image from the current timestep. We visualize the learned gating masks $\bm{M_{\Psi}}$ to better interpret the learned feature selection process. \figref{fig:gate} shows the visualizations from this experiment. For each input pair of images of height $H$ and width $W$, the gating branch outputs a single-channel mask of spatial size $\frac{H}{8} \times \frac{W}{8}$. In order to visualize the gating masks together with the input and output images, we first upsample the gating mask to $H\times W$ using Lanczos interpolation of order 4 over $8\times 8$ pixel neighborhood. Subsequently, we employ the jet color map on the masks. Values close to $0$ visualized with colors closer to blue indicate features that are selected from the inpainted image from the previous timestep, while values close to $1.0$ which are visualized with colors closer to red indicate the features that are selected from the coarsely inpainted image from the current timestep.

\figref{fig:gate}~(a) shows that the network uses the inpainted image from the previous timestep to suppress the shadow of the car and the reflection on the wet road while taking other parts of image almost exclusively from the coarsely inpainted image from the current timestep. Similarly, \figref{fig:gate}~(b) and \figref{fig:gate}~(c) also demonstrate that the shadows are removed using information from the inpainted image from the previous timestep and most other regions are taken from the coarsely inpainted image from the current timestep. We can also observe that for the areas that are occluded in the inpainted image from the previous timestep (e.g light pole on the right in \figref{fig:gate}~(b)), the network uses most of the features from the coarsely inpainted image from the current timestep (indicated by red in $\bm{M_{\Psi}}$). Furthermore, \figref{fig:gate}~(c) shows a scene without strong contrast and brightness discontinuities between the inpainted image from the previous timestep and the coarsely inpainted image from the current timestep. We observe that in this case the network uses the same amount of information from both images (indicated by green in $\bm{M_{\Psi}}$) to refine the target regions.

\begin{figure}
\centering
\footnotesize
\setlength{\tabcolsep}{0.05cm}
{\renewcommand{\arraystretch}{0.25}
\begin{tabular}{P{0.3cm}P{2cm}P{2cm}P{2cm}P{2cm}}
& \raisebox{-0.4\height}{Input+Semantics} & \raisebox{-0.4\height}{Empty Cities~\cite{bescos2019empty}} & \raisebox{-0.4\height}{DeepFill v2~\cite{yu2018free}} & \raisebox{-0.4\height}{DynaFill (Ours)} \\
\\
(a) & \raisebox{-0.4\height}{\includegraphics[width=\linewidth]{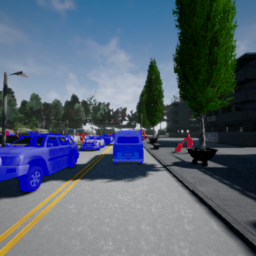}} & \raisebox{-0.4\height}{\includegraphics[width=\linewidth]{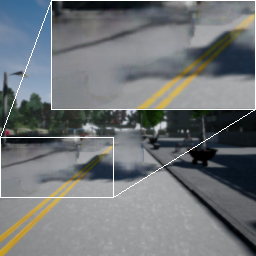}} & \raisebox{-0.4\height}{\includegraphics[width=\linewidth]{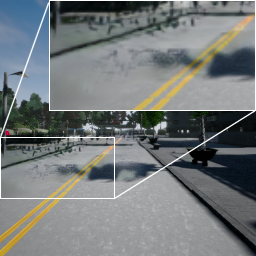}} & \raisebox{-0.4\height}{\includegraphics[width=\linewidth]{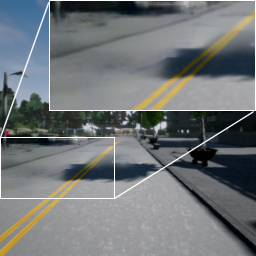}} \\
\\
(b) & \raisebox{-0.4\height}{\includegraphics[width=\linewidth]{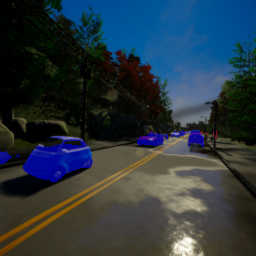}} & \raisebox{-0.4\height}{\includegraphics[width=\linewidth]{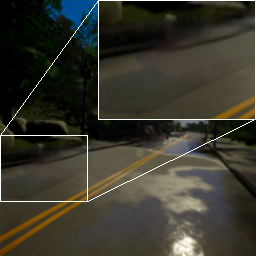}} & \raisebox{-0.4\height}{\includegraphics[width=\linewidth]{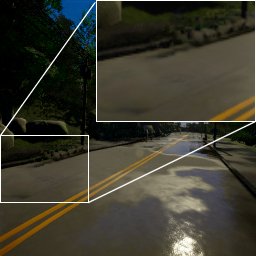}} & \raisebox{-0.4\height}{\includegraphics[width=\linewidth]{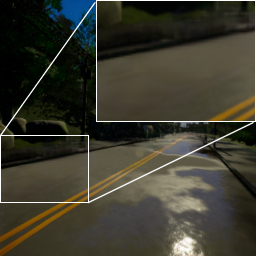}} \\
\\
(c) & \raisebox{-0.4\height}{\includegraphics[width=\linewidth]{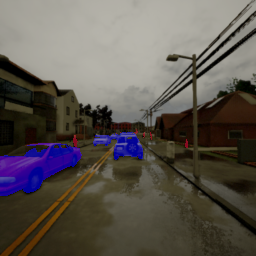}} & \raisebox{-0.4\height}{\includegraphics[width=\linewidth]{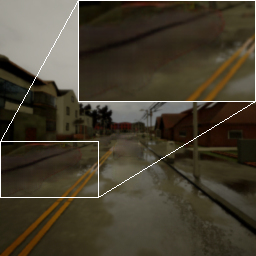}} & \raisebox{-0.4\height}{\includegraphics[width=\linewidth]{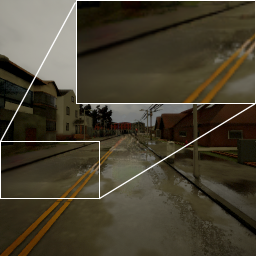}} & \raisebox{-0.4\height}{\includegraphics[width=\linewidth]{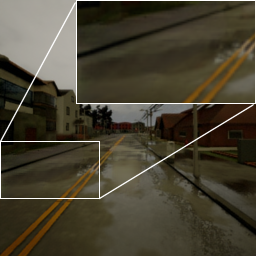}}
\end{tabular}}
\caption{Qualitative comparison with previous state-of-the-art methods on the validation set. We highlight the results by zooming in on parts of the scene occluded by dynamic objects in the input image.}
\label{fig:qualitative}
\end{figure}

\subsection{Qualitative Evaluations}
\label{sec:qualitative}

\begin{figure}
\centering
\footnotesize
\setlength{\tabcolsep}{0.05cm}
{\renewcommand{\arraystretch}{0.25}
\begin{tabular}{P{0.3cm}P{2cm}P{2cm}P{2cm}P{2cm}}
& \raisebox{-0.4\height}{RGB Input} & \raisebox{-0.4\height}{Depth Input} & \raisebox{-0.4\height}{DynaFill Output} & \raisebox{-0.4\height}{DynaFill Output} \\
\\
(a) & \raisebox{-0.4\height}{\includegraphics[width=\linewidth]{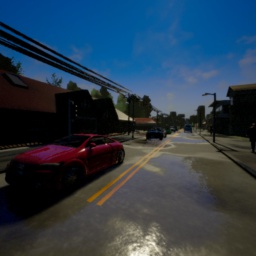}} & \raisebox{-0.4\height}{\includegraphics[width=\linewidth]{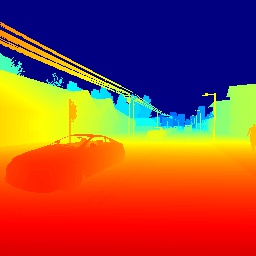}} & \raisebox{-0.4\height}{\includegraphics[width=\linewidth]{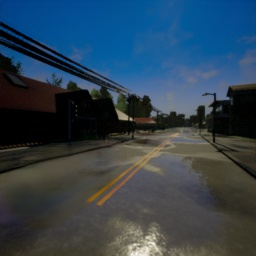}} & \raisebox{-0.4\height}{\includegraphics[width=\linewidth]{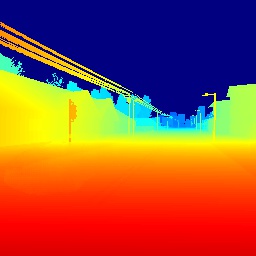}} \\
\\
(b) & \raisebox{-0.4\height}{\includegraphics[width=\linewidth]{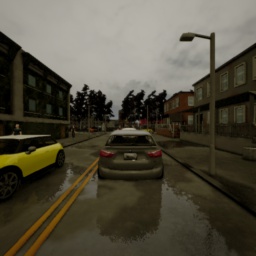}} & \raisebox{-0.4\height}{\includegraphics[width=\linewidth]{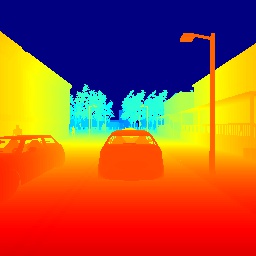}} & \raisebox{-0.4\height}{\includegraphics[width=\linewidth]{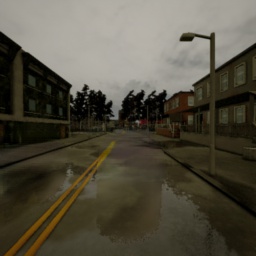}} & \raisebox{-0.4\height}{\includegraphics[width=\linewidth]{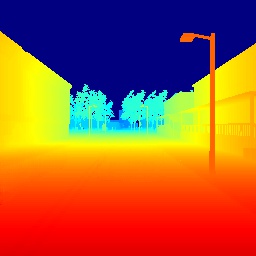}} \\
\\
(c) &
\raisebox{-0.4\height}{\includegraphics[width=\linewidth]{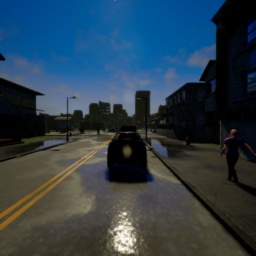}} & \raisebox{-0.4\height}{\includegraphics[width=\linewidth]{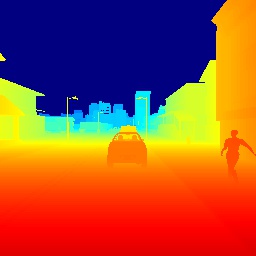}} & \raisebox{-0.4\height}{\includegraphics[width=\linewidth]{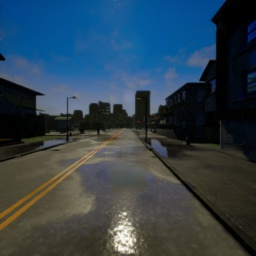}} & \raisebox{-0.4\height}{\includegraphics[width=\linewidth]{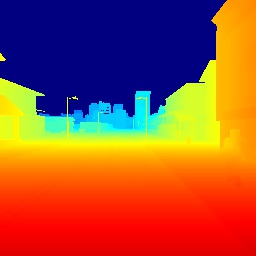}}
\end{tabular}}
\caption{Qualitative results of DynaFill. By conditioning depth completion on the inpainted image and incorporating the previously inpainted depth map in recurrent gated feedback, our model yields geometrically consistent results.}
\label{fig:qualitative_depth}
\end{figure}

\begin{figure}
\centering
\footnotesize
\setlength{\tabcolsep}{0.05cm}
{\renewcommand{\arraystretch}{0.25}
\begin{tabular}{P{0.3cm}P{2cm}P{2cm}P{2cm}P{2cm}}
& (a) & (b) & (c) & (d) \\
\\
\raisebox{0.2\height}{\rotatebox[origin=c]{90}{Input}} & \raisebox{-0.4\height}{\includegraphics[width=\linewidth]{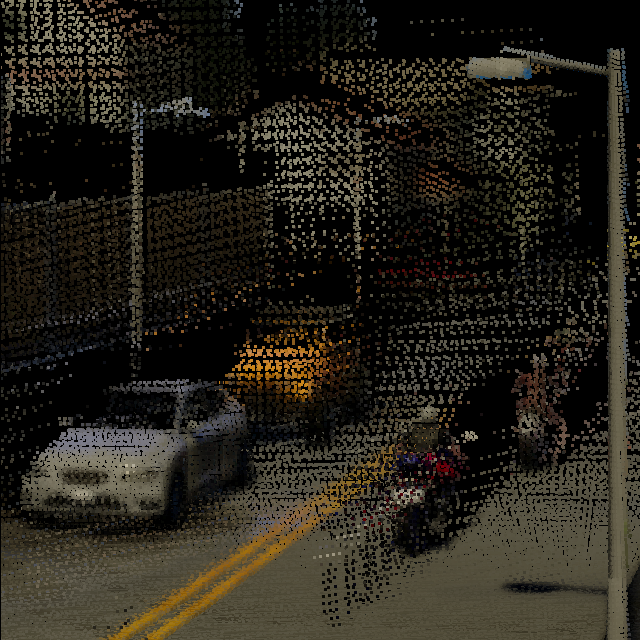}} &
\raisebox{-0.4\height}{\includegraphics[width=\linewidth]{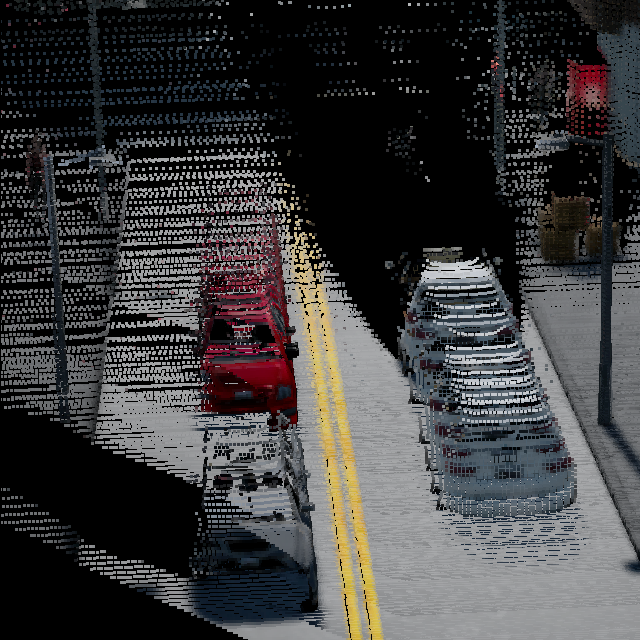}} &
\raisebox{-0.4\height}{\includegraphics[width=\linewidth]{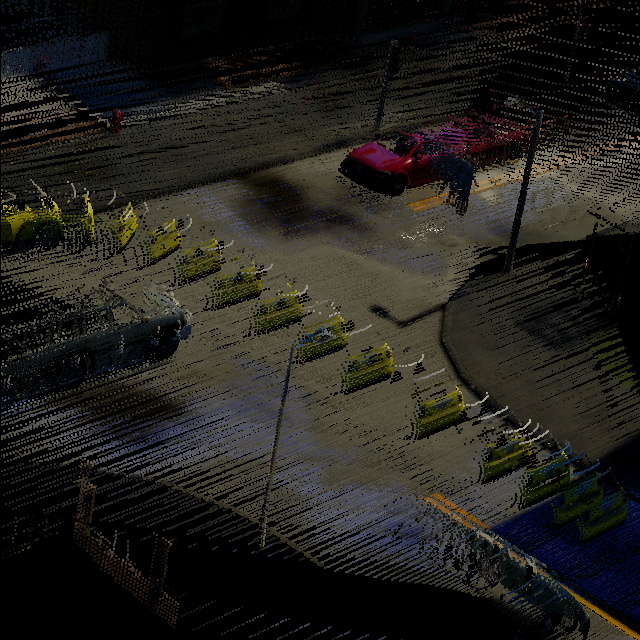}} &
\raisebox{-0.4\height}{\includegraphics[width=\linewidth]{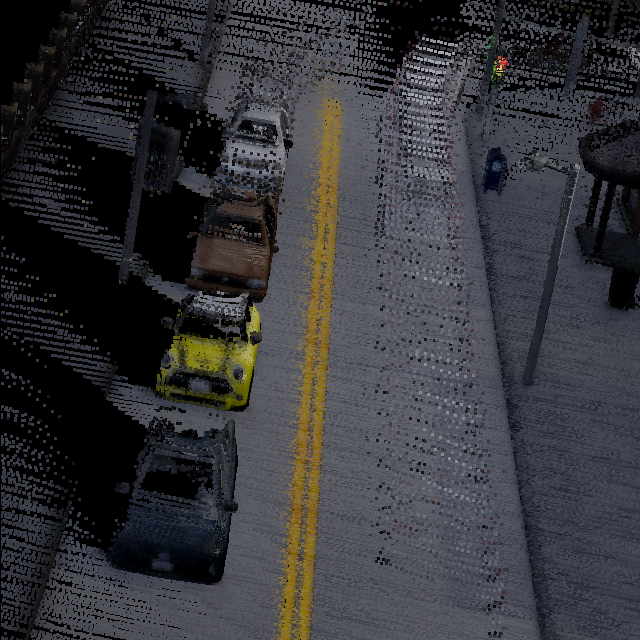}} \\
\\
\raisebox{0.2\height}{\rotatebox[origin=c]{90}{Semantics Output}} & 
\raisebox{-0.4\height}{\includegraphics[width=\linewidth]{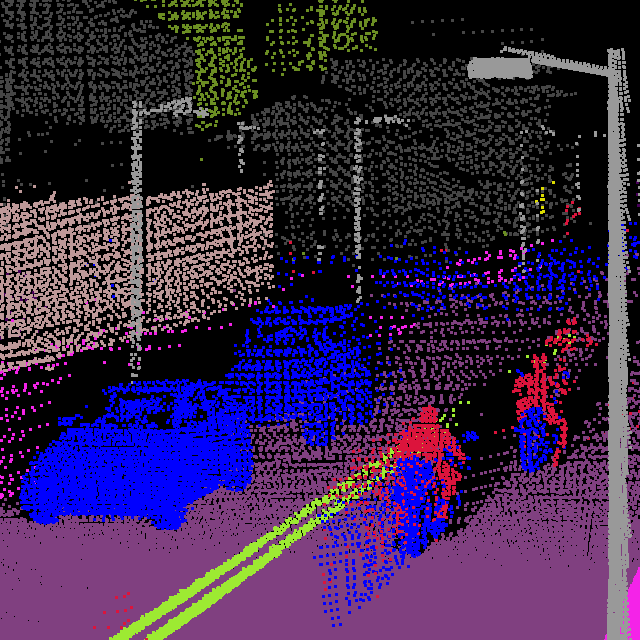}} &
\raisebox{-0.4\height}{\includegraphics[width=\linewidth]{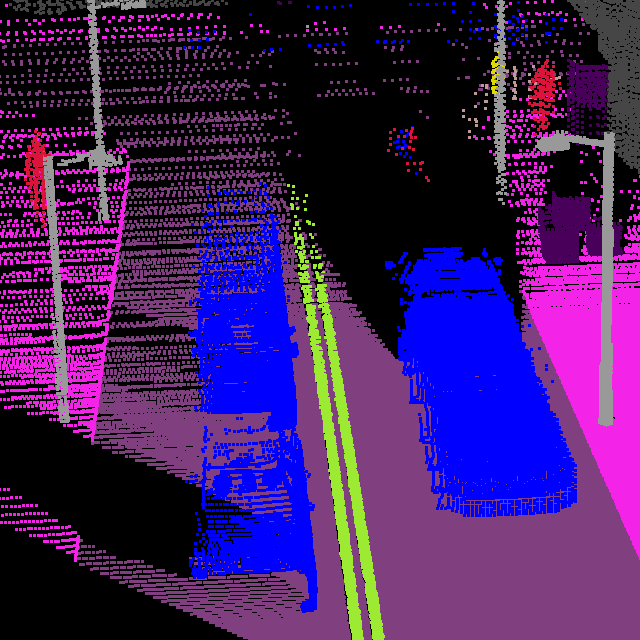}} &
\raisebox{-0.4\height}{\includegraphics[width=\linewidth]{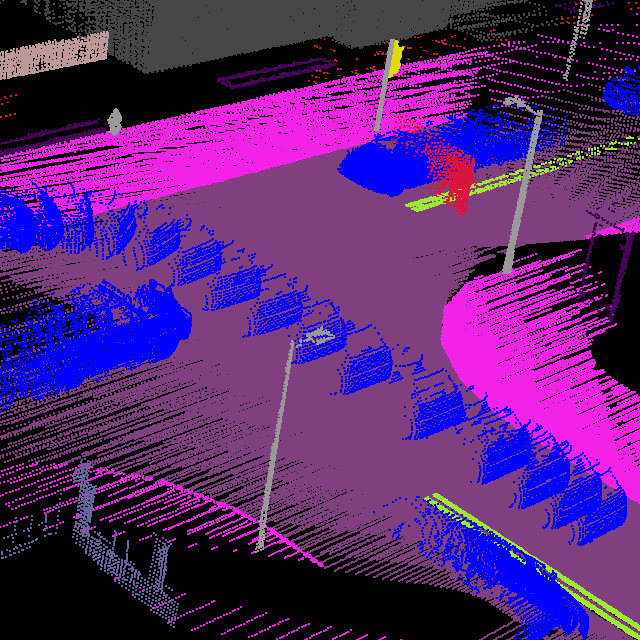}} &
 \raisebox{-0.4\height}{\includegraphics[width=\linewidth]{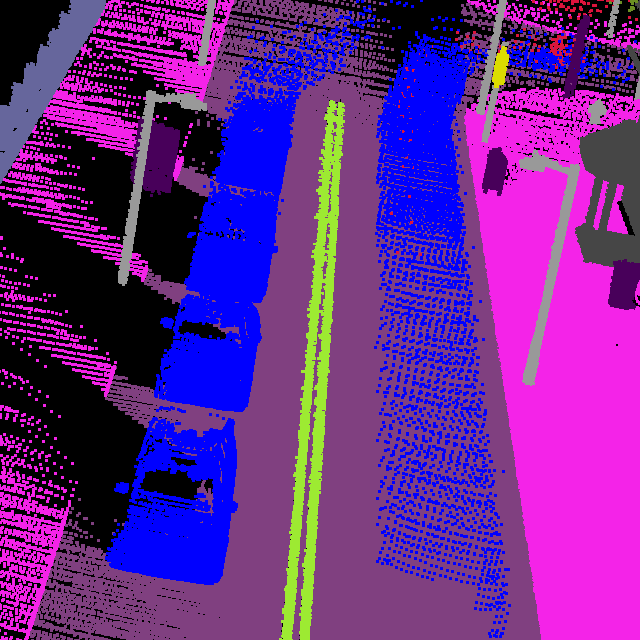}} \\
 \\
\raisebox{0.2\height}{\rotatebox[origin=c]{90}{DynaFill Output}} &   \raisebox{-0.4\height}{\includegraphics[width=\linewidth]{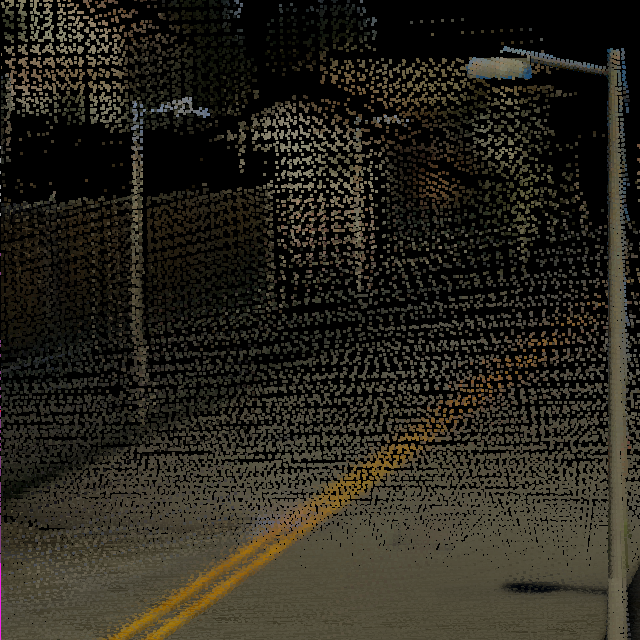}} &
\raisebox{-0.4\height}{\includegraphics[width=\linewidth]{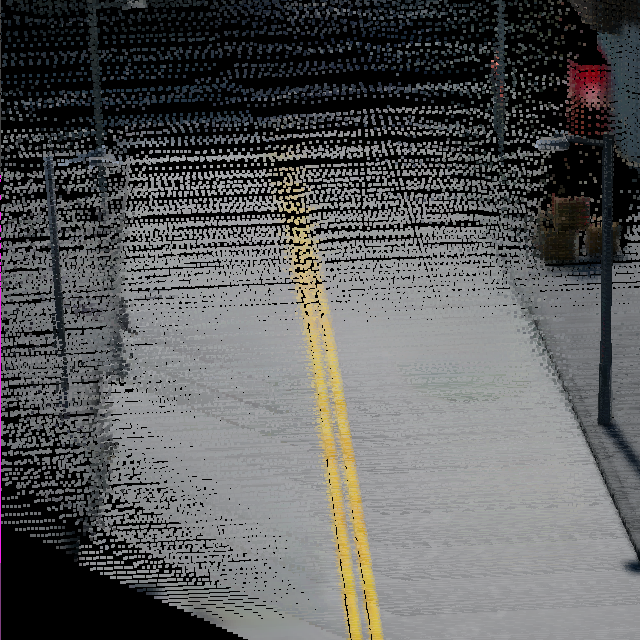}} &
\raisebox{-0.4\height}{\includegraphics[width=\linewidth]{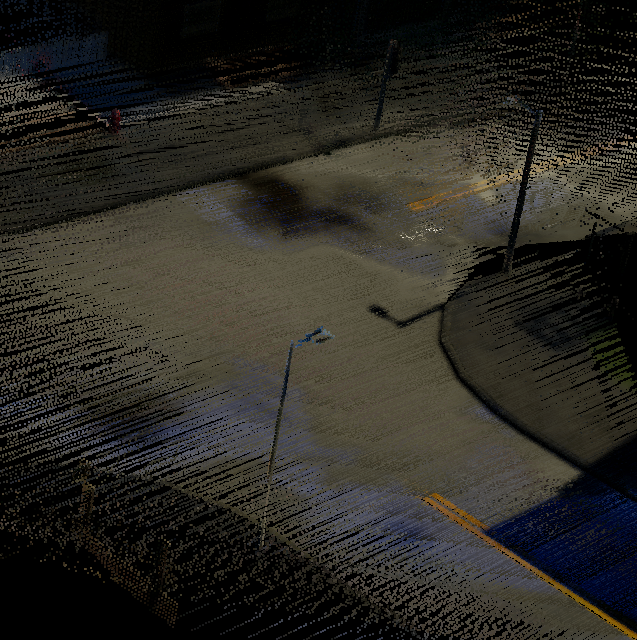}} &
 \raisebox{-0.4\height}{\includegraphics[width=\linewidth]{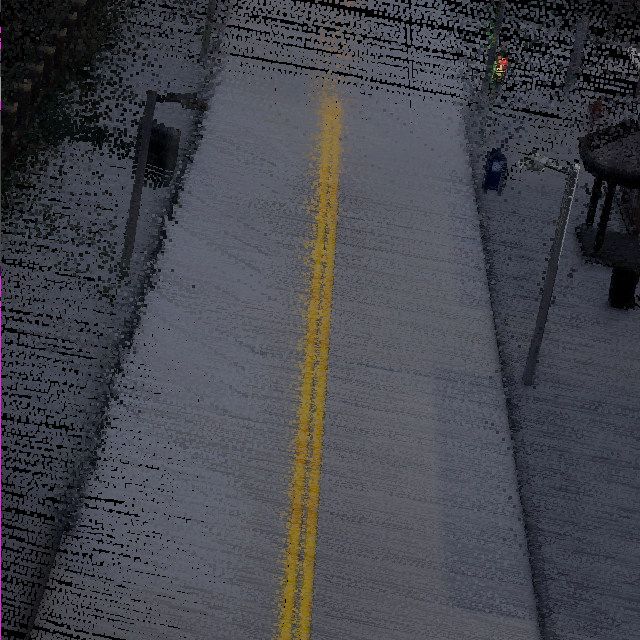}}
\end{tabular}}
\caption{Point cloud visualization of our RGB-D network outputs for multiple streams. DynaFill successfully recovers even bigger parts of the scene that are not visible over the entire duration of a trajectory due to being occluded by dynamic objects.}
\label{fig:qualitative_pcd}
\end{figure}

We qualitatively evaluate the performance of our proposed Dynafill model against the best performing previous state-of-the-art method DeepFill~v2 and Empty Cities in \figref{fig:qualitative}. We particularly study the hard cases for dynamic object removal where the objects are either close to the camera or in the image boundaries. We observe that DeepFill~v2 produces severe visual artifacts described by excessive and noisy patch replication while Empty Cities fails to completely remove the foreground objects and synthesizes blurry content in the target regions. DynaFill yields realistic colors and textures that are geometrically consistent with seamless boundary transitions as well as shadow/reflection removal. This can be attributed to effectively integrating both local spatial as well as temporal context by warping inpainted frames from the previous timesteps and fusing the feature maps using the learned mask in our recurrent gated feedback mechanism which enables adaptive reuse of information. Accurately inpainting occluded regions with information from the previous frames and inpainting them demonstrates that our network is able to reason about the geometry of the environment through inpainted depth maps. Additionally, we present qualitative results of the entire output of our network that contains both inpainted images and the corresponding inpainted depth maps in \figref{fig:qualitative_depth}. We can see that our network yields consistent results by regressing depth values in regions occluded by dynamic objects, conditioned on the inpainted images and by temporally warping inpainted images from previous timesteps using the inpainted depth maps. Moreover, to illustrate complex inpainting scenarios in urban driving scenes, we show the aggregated textured pointclouds in \figref{fig:qualitative_pcd}. Note that we do not perform scan matching, we only aggregate scans along a single trajectory. We observe that DynaFill is able to successfully recover large parts of the scene that are not visible over the entire duration of a trajectory and the results demonstrate color, geometric and temporal consistency in the inpainted outputs.

\begin{figure*}
\centering
\footnotesize
\setlength{\tabcolsep}{0.05cm}
{\renewcommand{\arraystretch}{0.3}
\begin{tabular}{P{0.5cm}P{2.5cm}P{2.5cm}P{2.5cm}P{2.5cm}P{2.5cm}}
& \raisebox{-0.4\height}{t - 4} & \raisebox{-0.4\height}{t - 3} & \raisebox{-0.4\height}{t - 2} & \raisebox{-0.4\height}{t - 1} & \raisebox{-0.4\height}{t} \\
\\
\raisebox{0.2\height}{\rotatebox[origin=c]{90}{Input}} &
\raisebox{-0.4\height}{\includegraphics[width=\linewidth]{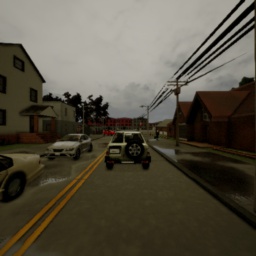}} & 
\raisebox{-0.4\height}{\includegraphics[width=\linewidth]{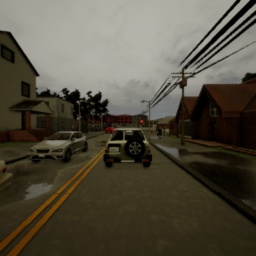}} & 
\raisebox{-0.4\height}{\includegraphics[width=\linewidth]{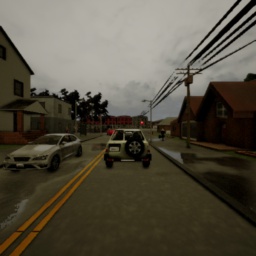}} & 
\raisebox{-0.4\height}{\includegraphics[width=\linewidth]{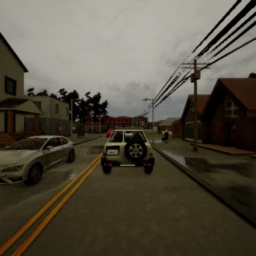}} &
\raisebox{-0.4\height}{\includegraphics[width=\linewidth]{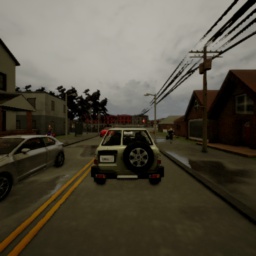}} \\
\\
\raisebox{0.2\height}{\rotatebox[origin=c]{90}{Empty Cities~\cite{bescos2019empty}}} &
\raisebox{-0.4\height}{\includegraphics[width=\linewidth]{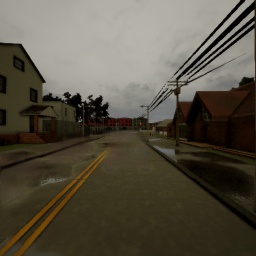}} & 
\raisebox{-0.4\height}{\includegraphics[width=\linewidth]{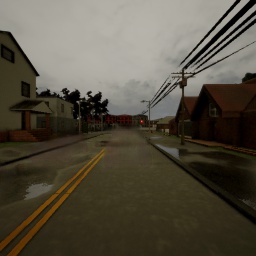}} & 
\raisebox{-0.4\height}{\includegraphics[width=\linewidth]{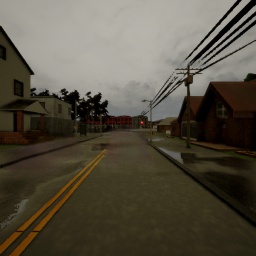}} & 
\raisebox{-0.4\height}{\includegraphics[width=\linewidth]{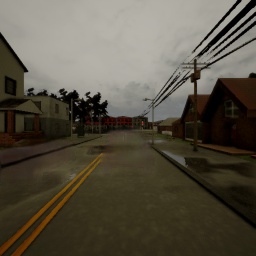}} &
\raisebox{-0.4\height}{\includegraphics[width=\linewidth]{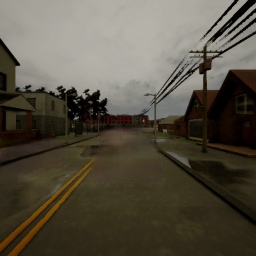}} \\
\\
\raisebox{0.2\height}{\rotatebox[origin=c]{90}{DeepFill~v2~\cite{yu2018free}}} &
\raisebox{-0.4\height}{\includegraphics[width=\linewidth]{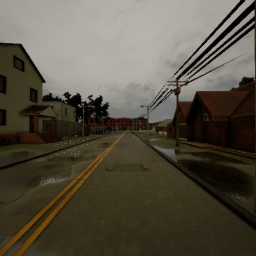}} & 
\raisebox{-0.4\height}{\includegraphics[width=\linewidth]{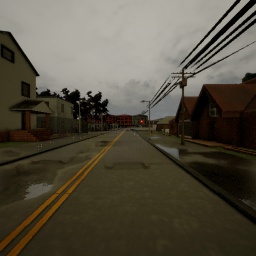}} & 
\raisebox{-0.4\height}{\includegraphics[width=\linewidth]{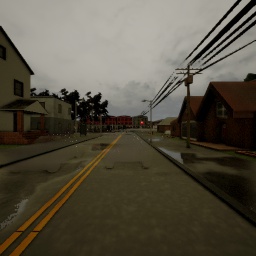}} & 
\raisebox{-0.4\height}{\includegraphics[width=\linewidth]{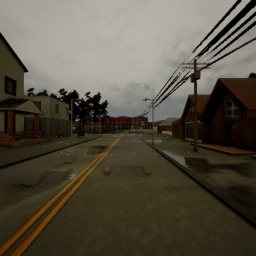}} &
\raisebox{-0.4\height}{\includegraphics[width=\linewidth]{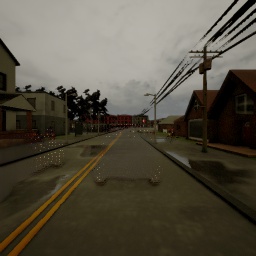}} \\
\\
\raisebox{0.2\height}{\rotatebox[origin=c]{90}{LGTSM~\cite{chang2019learnable}}} &
\raisebox{-0.4\height}{\includegraphics[width=\linewidth]{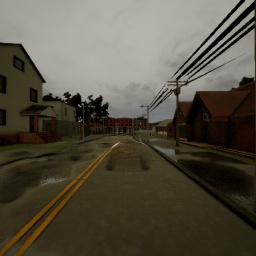}} & 
\raisebox{-0.4\height}{\includegraphics[width=\linewidth]{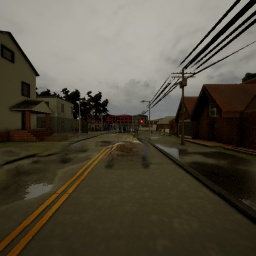}} & 
\raisebox{-0.4\height}{\includegraphics[width=\linewidth]{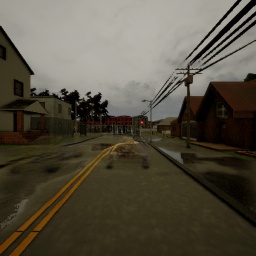}} & 
\raisebox{-0.4\height}{\includegraphics[width=\linewidth]{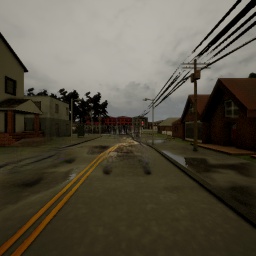}} &
\raisebox{-0.4\height}{\includegraphics[width=\linewidth]{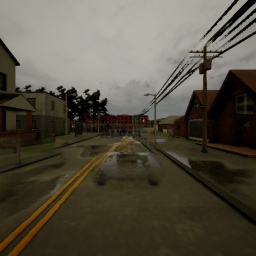}} \\
\\
\raisebox{0.2\height}{\rotatebox[origin=c]{90}{DynaFill (Ours)}} &
\raisebox{-0.4\height}{\includegraphics[width=\linewidth]{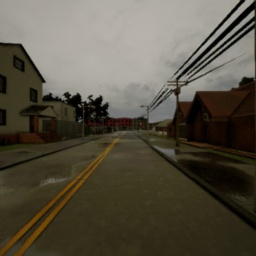}} & 
\raisebox{-0.4\height}{\includegraphics[width=\linewidth]{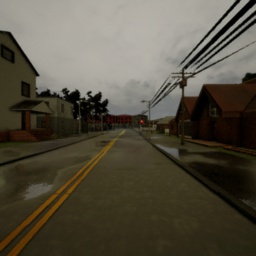}} & 
\raisebox{-0.4\height}{\includegraphics[width=\linewidth]{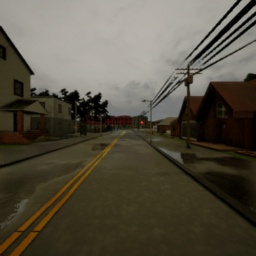}} & 
\raisebox{-0.4\height}{\includegraphics[width=\linewidth]{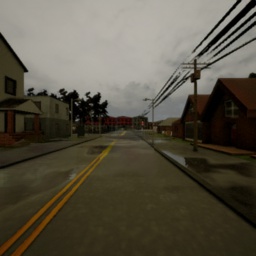}} &
\raisebox{-0.4\height}{\includegraphics[width=\linewidth]{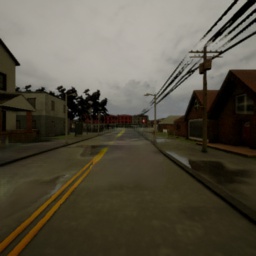}}
\end{tabular}}
\caption{Qualitative comparison of spatio-temporal consistency over sequence-18. Our DynaFill model hallucinates spatio-temporally consistent results while generating background details in regions occluded by the dynamic objects, such as lane markings.}
\label{fig:qualitative_temporal_b}
\end{figure*}

\begin{figure}
\centering
\footnotesize
\setlength{\tabcolsep}{0.05cm}
{\renewcommand{\arraystretch}{0.25}
\begin{tabular}{P{0.3cm}P{2cm}P{2cm}P{2cm}P{2cm}}
& \raisebox{-0.4\height}{Query (Raw)} & \raisebox{-0.4\height}{Result (Raw)} & \raisebox{-0.4\height}{Query (Ours)} & \raisebox{-0.4\height}{Result (Ours)} \\
\\
(a) &
\raisebox{-0.4\height}{\includegraphics[width=\linewidth]{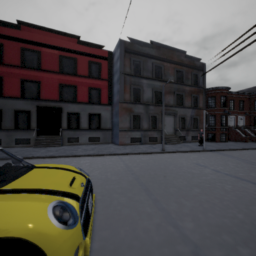}} & 
\raisebox{-0.4\height}{\includegraphics[width=\linewidth]{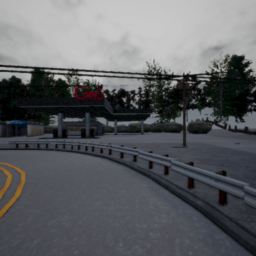}} & 
\raisebox{-0.4\height}{\includegraphics[width=\linewidth]{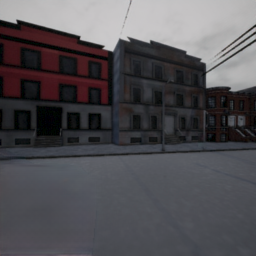}} & \raisebox{-0.4\height}{\includegraphics[width=\linewidth]{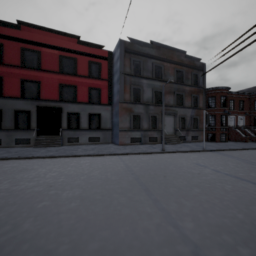}} \\
\\
(b) &
\raisebox{-0.4\height}{\includegraphics[width=\linewidth]{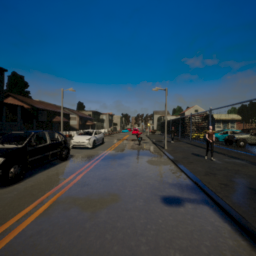}} & 
\raisebox{-0.4\height}{\includegraphics[width=\linewidth]{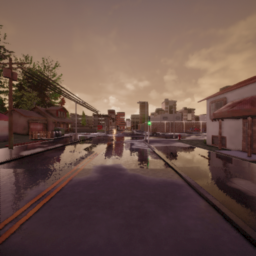}} & \raisebox{-0.4\height}{\includegraphics[width=\linewidth]{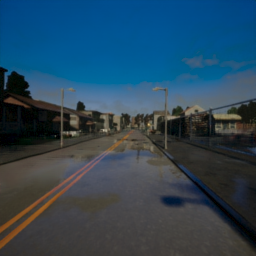}} & \raisebox{-0.4\height}{\includegraphics[width=\linewidth]{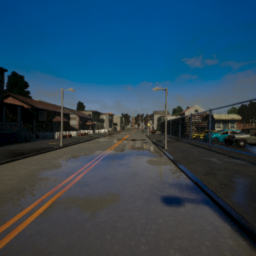}} \\
\\
(c) &
\raisebox{-0.4\height}{\includegraphics[width=\linewidth]{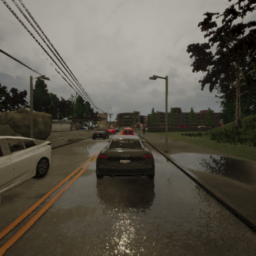}} & 
\raisebox{-0.4\height}{\includegraphics[width=\linewidth]{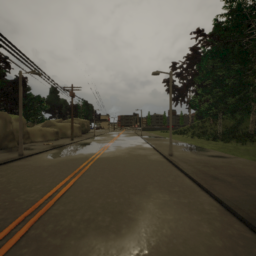}} & \raisebox{-0.4\height}{\includegraphics[width=\linewidth]{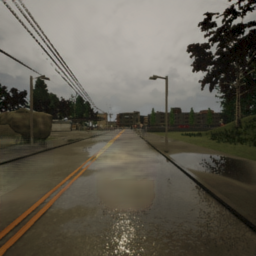}} & \raisebox{-0.4\height}{\includegraphics[width=\linewidth]{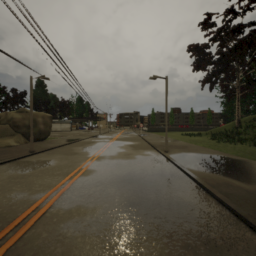}} \\
\end{tabular}}
\caption{Qualitative evaluation of DenseVLAD~\protect\cite{dense_vlad} retrieval results with DynaFill as a preprocessor. In contrast to the raw image query which often results in localization errors, removing dynamic objects and inpainting them with our model enables the system to retrieve accurate nearest neighbor frames.}
\label{fig:qualitative_densevlad}
\end{figure}

We qualitatively evaluate the spatio-temporal consistency achieved by DynaFill over sequences of 5 consecutive frames in \figref{fig:qualitative_temporal_b}. We compare against the state-of-the-art image-to-image translation-based dynamic object removal method (Empty Cities~\cite{bescos2019empty}), image inpainting (DeepFill~v2~\cite{yu2019free}) and video inpainting models (LGTSM~\cite{chang2019learnable}). While DeepFill~v2 typically performs well at inpainting target regions, it yields visually unappealing results due to its inability to remove any effects induced by dynamic objects such as shadows or reflections. Moreover, it suffers from noisy patch replication which produces flickering between neighboring frames, especially in regions with frequent brightness changes. This leads to temporally inconsistent results. LGTSM being a video inpainting model, yields inconsistent results when there is large motion between frames despite the fact it uses the future frame information. Although it often maintains temporal consistency, large motions cause noisy results in the spatial domain and we also observe that it replicates shadows induced by vehicles, into neighboring inpainted frames. Finally, Empty Cities performs dynamic-to-static image translation aimed at removing artifacts induced by dynamic objects such as shadows or reflections, together with inpainting dynamic objects regions. We observe that the hallucinated content inside the masked regions is often over-smoothed with inaccurate color (\textit{e.g.} in saturation) and geometrical inconsistencies (\textit{e.g.} wavy curbs) between frames which causes prominent silhouettes of dynamic objects that are visible over time. In contrast to these methods, our proposed DynaFill model accurately removes dynamic objects and corrects correlated regions containing shadows or reflections, while being spatio-temporally consistent and robust to illumination changes. It only uses the previous inpainted frame information in a recurrent manner, thereby it directly handles large motion between consecutive frames and the inference time makes it suitable for real-time applications. Additionally, videos demonstrating the spatio-temporal consistency achieved by our DynaFill model is shown in \url{http://rl.uni-freiburg.de/research/rgbd-inpainting}.

While we have evaluated our framework on highly realistic synthetic data, it is important to discuss the feasibility of training such models directly on real-world data. A straightforward approach would be to collect a real-world dataset of RGB-D videos with odometry information and groundtruth containing the background in the regions occluded by dynamic objects and train our framework on it. A more promising approach to mitigate the need for groundtruth without dynamic objects would be to employ cycle-consistent adversarial networks or self-supervised approaches, such as training the network on real-world images with randomly generated masks. Another promising avenue would be to employ unsupervised domain adaptation techniques on our trained model. Investigating these avenues would bring us closer to having an effective solution in the real world.


\subsection{Retrieval-based Localization}
\label{sec:localization}

\begin{table}
\footnotesize
\centering
\caption{DenseVLAD~\cite{dense_vlad} query accuracies for different inputs (in \%).}
\label{dense_vlad}
\setlength{\tabcolsep}{0.15cm}
\begin{tabular}{p{2cm}|P{0.7cm}P{0.7cm}|P{1cm}P{1.1cm}P{1.2cm}}
\toprule
 &  &  & \multicolumn{3}{c}{Threshold} \\
Query input & Top-1 & Top-5 & \SI{5}{\meter}, \SI{10}{\degree} & \SI{0.5}{\meter}, \SI{5}{\degree} & \SI{0.25}{\meter}, \SI{2}{\degree} \\
\noalign{\smallskip}\hline\hline\noalign{\smallskip}
Raw & 62.45 & 86.62 & 79.29 & 68.16 & 67.08 \\
\textbf{Inpainted} & \textbf{82.30} & 95.87 & 92.13 & 86.53 & 85.35 \\
\textbf{Image-to-image} & 81.91 & \textbf{96.95} & \textbf{93.51} & \textbf{87.22} & \textbf{85.94} \\
\bottomrule
\end{tabular}
\end{table}

One of the use cases of our temporal RGB-D inpainting framework is for visual localization tasks. In order to demonstrate the potential benefits, we performed experiments by employing our model as a preprocessor for retrieval-based visual localization on the entire validation set of our dataset.
We use DenseVLAD~\cite{dense_vlad} due to its simplicity and for the fact of being one of the state-of-the-art methods.
In our experiments, we use $N = 25 \cdot {10}^6$ (number of descriptors), $k = 128$ (number of visual words), $d = 8$ (final PCA dimensionality). The descriptors were selected randomly using reservoir sampling.

Using the benchmarking evaluation protocol~\cite{visual_localization_benchmark}, we report the percentage of query images with predicted 6-DoF poses that are within three error tolerance thresholds of $(\SI{5}{\meter},\SI{10}{\degree})$, $(\SI{0.5}{\meter},\SI{5}{\degree})$, and $(\SI{0.25}{\meter},\SI{2}{\degree})$. For the sake of completeness, we also report Top-1 and Top-5 accuracies. \tabref{dense_vlad} shows the results from this experiment. Our inpainting model achieves an improvement of $12.84\%, 18.37\%$ and $18.27\%$ over the non-inpainted image, across all the three thresholds respectively. While our image-to-image translation model further improves the performance over the non-inpainted image by $14.2\%$, $19.1\%$ and $18.9\%$ respectively. 

We further show qualitative retrieval results with different dynamic object configurations, weather conditions and times of day in \figref{fig:qualitative_densevlad} by comparing with the raw input image query as a baseline and our DynaFill model as a query preprocessor. The variance in the extracted descriptors, induced by the movement of dynamic objects through the scene causes the system to retrieve frames with substantial pose error as shown in the quantitative results in \tabref{dense_vlad}. We observe that in the qualitative retrieval results shown in \figref{fig:qualitative_densevlad}~(a), the yellow color of the car in the raw image query produces an outlier in the DenseVLAD descriptor. This causes the localization system to retrieve a frame that has a similar yellow curve in the lower left part of the image, \textit{i.e.} a yellow lane marking. Sensitivity to specific features in an image, \textit{e.g.} reflections on a wet road or very bright colors, may lead the system to retrieve frames that were recorded at a different time of the day and from a completely different location. This is best illustrated by \figref{fig:qualitative_densevlad}~(b) in which the query result is a large statistical outlier. \figref{fig:qualitative_densevlad}~(c) shows an example where the system is able to recognize the approximate area, however irrelevant descriptor features that are induced by cars overpower the informative ones causing significant localization errors. More importantly, we observe that by employing our DynaFill model as a preprocessor that takes the raw images with dynamic objects as input and inpaints dynamic object regions with background content, it suppresses the outliers produced by DenseVLAD and enables us to accurately localize. We obtain consistently accurate retrieval results in all the examples shown in \figref{fig:qualitative_densevlad}. These results demonstrate substantial improvements that can be achieved using our model as an out-of-the-box solution for removing dynamic objects for localization and mapping systems.

\subsection{Limitations and Future Work}

In this section, we discuss the limitations of our approach and provide directions for future work. Our approach assumes a paired RGB-D image data is available. Obtaining such dynamic/static pairs of real-world data with groundtruth pixel-level labels of dynamic objects is still an extremely arduous task. Moreover, our method relies on forward warping of previous frames that improves the performance of the inpainting model, but also introduces the sensitivity to inaccurate estimation of the odometry and depth inpainting. The quality of inpainting also depends of the quality of the dynamic mask, which in some cases such as low visibility, can be hard to estimate. Finally, we operate inside an adversarial framework that comes with its own downsides. Cumbersome selection of hyperparameters and long and unstable training processes are just some of the side effects that we have experienced.

As future work, the joint inpainting of different modalities can be investigated. That is, inpainting RGB frames together with their corresponding semantic segmentation can be crutial in achieving higher levels of spatial consistency. Moreover, employing odometry for additional forward warping of semantics and depth can potentially improve the temporal performance of our method over multiple time steps. Instead of only localization, a SLAM system could be introduced into both training and testing phases of our system to provide useful information, with additional constraints guiding the optimization process and making the training more stable. Such an approach would incorporate feedback from a built map into the inpainting process, by exploiting global poses available in our newly introduced large-scale dataset. Additionally, our dataset shows a large potential for studying the multimodal cyclic consistency in generative adversarial networks, applicable to both inpainting and various other related tasks. One of the ideas is also to inpaint random parts of images that do not belong to dynamic objects. This would mean casting the inpainting as a self-supervised task and is be an important step towards mitigating the need for such a complex dataset of paired RGB-D image data.

\section{Conclusions}
\label{sec:conclusions}

In this paper, we proposed an end-to-end deep learning architecture for dynamic object removal and inpainting from temporal RGB-D sequences. Our coarse-to-fine model trained under the generative adversarial framework synthesizes spatially coherent realistic color as well as textures and enforces temporal consistency using a gated recurrent feedback mechanism that adaptively fuses information from previously inpainted frames using odometry and the previously inpainted depth map. Our model encourages geometric consistency during end-to-end training of the our inpainting architecture by conditioning the depth completion on the inpainted image and simultaneously using the previously inpainted depth map in the feedback mechanism. As opposed to existing video inpainting methods, our model does not utilize future frame information and produces more accurate and visually appealing results by also removing shadows or reflections from regions surrounding dynamic objects.

We introduced a large-scale hyperealistic dataset with \mbox{RGB-D} sequences and groundtruth information of occluded regions that we have made publicly available. We performed extensive experiments that show that our DynaFill model exceeds the performance of state-of-the-art image and video inpainting methods with a runtime suitable for real-time applications ($\sim 20$ FPS). Additionally, we presented detailed ablation studies, qualitative analysis and visualizations that highlight the improvement brought about by various components of our architecture. Furthermore, we presented experiments by employing our model as a preprocessor for retrieval-based visual localization that demonstrates the utility of our approach as an out-of-the-box front end for localization and mapping systems.


\section*{Acknowledgements}

This work was partly funded by the European Union’s Horizon 2020 research and innovation program under grant agreement No 871449-OpenDR and a research grant from Eva Mayr-Stihl Stiftung.

\ifCLASSOPTIONcaptionsoff
  \newpage
\fi


\bibliographystyle{IEEEtran}
\bibliography{besic21t-iv.bib}

\begin{IEEEbiography}[{\includegraphics[width=1in,height=1.25in,clip,keepaspectratio]{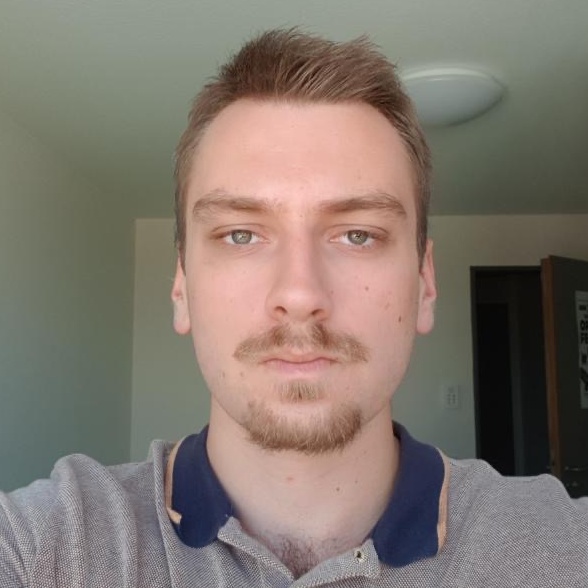}}]{Borna Be\v{s}i\'{c}} is a master's student in the Robot Learning Lab headed by Abhinav Valada. He received his bachelor's degree in Computer Science from University of Zagreb, Croatia. 
\end{IEEEbiography}

\begin{IEEEbiography}[{\includegraphics[width=1in,height=1.25in,clip,keepaspectratio]{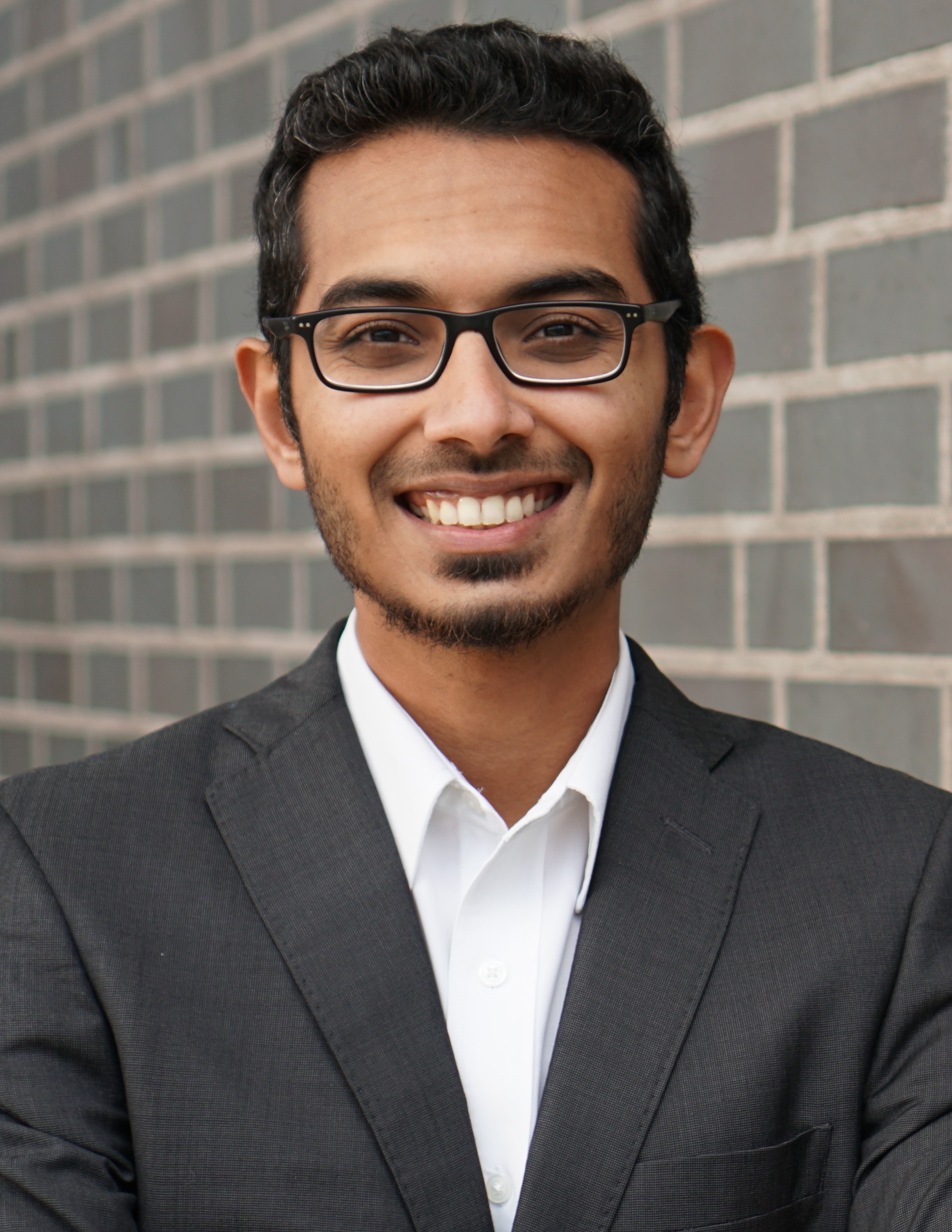}}]{Abhinav Valada}
is an Assistant Professor and Director of the Robot Learning Lab at the University of Freiburg. He is a member of the Department of Computer Science, a principal investigator at the BrainLinks-BrainTools Center, and a founding faculty of the European Laboratory for Learning and Intelligent Systems (ELLIS) unit at Freiburg. He received his Ph.D.~in Computer Science from the University of Freiburg in 2019 and his M.S.~degree in Robotics from Carnegie Mellon University in 2013. His research lies at the intersection of robotics, machine learning and computer vision with a focus on tackling fundamental robot perception, state estimation and control problems using learning approaches in order to enable robots to reliably operate in complex and diverse domains. Abhinav Valada is a Scholar of the ELLIS Society, a DFG Emmy Noether Fellow, and co-chair of the IEEE RAS TC on Robot Learning.
\end{IEEEbiography}


\vfill


\end{document}